\documentclass[runningheads]{llncs}

\usepackage{graphicx}

\usepackage{hyperref}

\usepackage{amsmath, amssymb}
\usepackage{subcaption}

\newcommand{\papertitle}{Handwriting Recognition with Novelty}
\newcommand{\papertitleshort}{\papertitle}

\begin{document}

\title{
    \papertitle\thanks{This research was sponsored by the Defense Advanced Research Projects Agency (DARPA) and the Army Research Office (ARO) under multiple contracts/agreements including HR001120C0055, W911NF-20-2-0005,W911NF-20-2- 0004,HQ0034-19-D-0001, W911NF2020009. The views contained in this document are those of the authors and should not be interpreted as representing the official policies, either expressed or implied, of the DARPA or ARO, or the U.S. Government.}
}

\titlerunning{\papertitleshort}

\author{
    Derek S. Prijatelj\inst{1}\orcidID{0000-0002-0529-9190}
    \and Samuel Grieggs\inst{1}\orcidID{0000-0002-2433-5257}
    \and Futoshi Yumoto\inst{2}\orcidID{0000-0001-5354-8254}
    \and Eric Robertson\inst{2}\orcidID{0000-0002-4942-4619}
    \and Walter J. Scheirer\inst{1}\orcidID{0000-0001-9649-8074}
}

\authorrunning{D. Prijatelj et al.}

\institute{
        University of Notre Dame, Notre Dame IN 46556, USA\\
        \email{\{dprijate, sgrieggs, walter.scheirer\}@nd.edu}
    \and
        PAR Government, 421 Ridge St, Rome NY 13440, USA\\
        \email{\{eric\_robertson, futoshi\_yumoto\}@partech.com}
}

\maketitle

\begin{abstract}

    This paper introduces an agent-centric approach to handle novelty in the visual recognition domain of handwriting recognition (HWR).
An ideal transcription agent would rival or surpass human perception, being able to recognize known and new characters in an image, and detect any stylistic changes that may occur within or across documents.
A key confound is the presence of novelty, which has continued to stymie even the best machine learning-based algorithms for these tasks.
In handwritten documents, novelty can be a change in writer,  character attributes, writing attributes, or overall document appearance, among other things.
Instead of looking at each aspect independently, we suggest that an integrated agent that can process known characters and novelties simultaneously is a better strategy.
This paper formalizes the domain of handwriting recognition with novelty, describes a baseline agent, introduces an evaluation protocol with benchmark data, and provides experimentation to set the state-of-the-art.
Results show feasibility for the agent-centric approach, but  more work is needed to approach human-levels of reading ability, giving the HWR community a formal basis to build upon as they solve this challenging problem.
\end{abstract}

\section{Introduction}

Reading comprehension is a complex human activity that requires symbol acquisition and manipulation, the perception of salient information, and an understanding of what is known and what is novel on a page~\cite{rayner2012psychology}.
Why has it been reduced to simple optical character recognition (OCR) within the field of machine learning?
While this has decreased the complexity of the domain to make it more tractable for standard data-driven approaches, it has also steered researchers away from the core of one of the most important competencies of natural intelligence: handling novelty.
Consequently, OCR algorithms are not effective on handwritten documents that exhibit a wide degree of variation in appearance~\cite{smith2018}.
Yet such documents can be effortlessly read by humans, even children who have recently become literate.
Fundamentally, this task is made difficult by the presence of novelties that are unknown at training time, which must be expected when a writer can do essentially whatever they want on a page.
In this paper we introduce an agent-centric approach to the handwriting recognition (HWR) domain with novelty to address this challenge. 

Novelty, of course, is not unique to the HWR domain.
In general, the ability to act appropriately and effectively in novel situations that occur in open worlds, as opposed to closed datasets, has been singled out as a crucial challenge in AI that is inhibiting progress in multiple domains~\cite{Senator19}.
Recent theoretical work has sought to understand what implications the unknown has in the context of activity and perceptual domains, and how it should be treated by an agent operating within them.
This includes risk formulations that account for the unknown~\cite{scheirer2012toward,fei2016breaking}, generative models of novelty~\cite{langley2020open}, and enumerations of the different possible types of novelty that can occur in practice~\cite{boult2020unifying}. 

As Boult \textit{et al.}~have noted, a universal theory of novelty is not possible to construct. This is because each domain will require its own definitions for a world state, dissimilarity (\textit{i.e.}, how far away from known data something must be to be considered novel), regret (\textit{i.e.}, the consequence of not detecting novelty) and other constructs needed to solve a given task within a specific domain.
Accordingly, researchers must formalize and attempt to solve tasks within a variety of domains to test the generalization capabilities of core novelty processing algorithms.
Handwritten documents are particularly well suited for the study of novelty because of the human creativity that goes into making them. Thus we propose this domain as a challenge problem for researchers not only studying HWR, but also novelty in AI. 

\begin{figure}[t]
    \centering
    \includegraphics[width=1.0\textwidth]{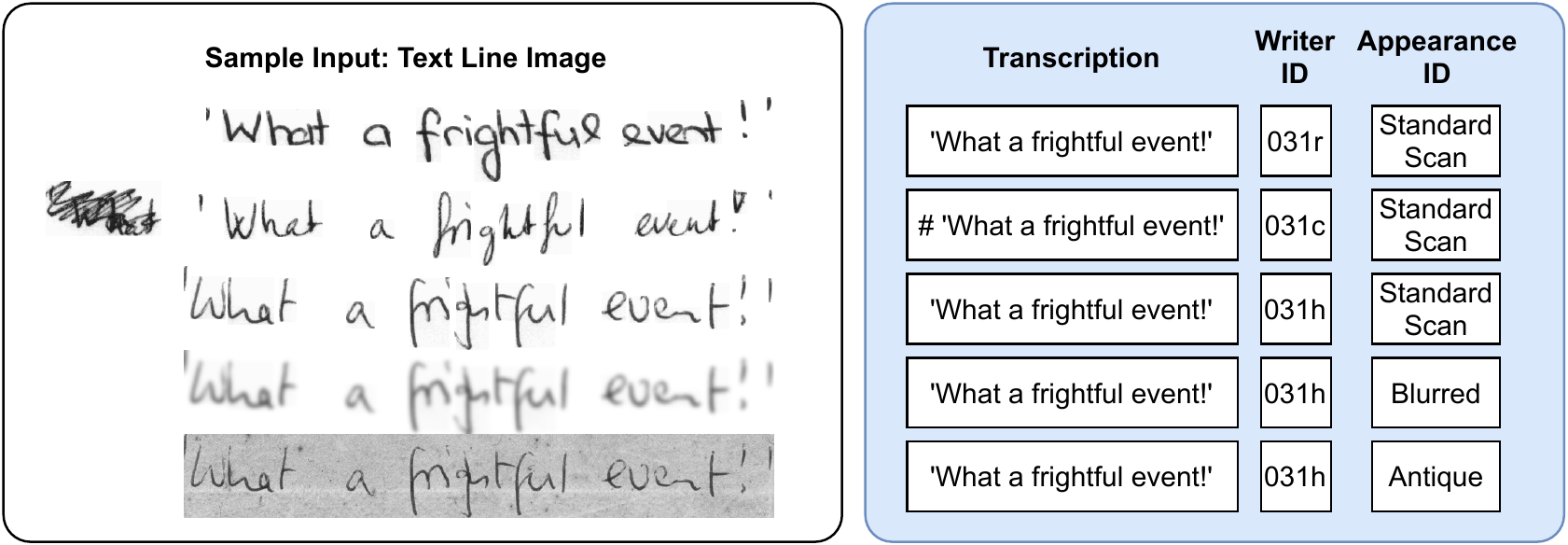}
    \caption{
        Examples from the IAM offline handwriting dataset~\cite{marti2002iam} depicting the difference in writing style between writers, unknown characters in the scratched out entry represented as ``\#" in the transcription, and the difference in appearance between the same samples with new modifications to simulate real-world situations.
        Novelty may occur in any of these labels, such as novel characters, writer, or appearance.
        HWR agents should be able to handle such novelty.
    }
    \label{fig:teaser}
\end{figure}

Beyond new characters, novelties may occur in the stylistic attributes of the written text.
The discriminating markers of a writer's style are the (often subtle) differences in the way characters are drawn, as seen in Fig.~\ref{fig:teaser}.
Global changes to the appearance of an image are also important, such as changes to the background or the application of filters (\textit{e.g}, Photoshop's vintage photo filter), which may have been made after a page has been acquired to improve human readability or to stylize to match the overall look of a digital text edition.
Not only are these elements important for managing novelty in ordinary transcription tasks, but they are also important to scholars who are interested in identifying stylistic markers of historical significance~\cite{stutzmann2016clustering,van2019paleography}. 

An agent-centric approach reflects the real human behavior of reading.
Individual algorithms can be applied to specific tasks in HWR, such as transcription or visual style recognition, but information can't easily be passed between them, nor can it be used in an informed joint manner.
Here we suggest that an agent with integrated task-specific modules and novelty-specific modules can autonomously process multiple information streams, understanding what it has seen before and what it hasn't, while it performs a transcription task.
Further, the novelty-specific modules should not exclusively rely on simple thresholding applied over standard classifiers, and instead consist of open world classifiers optimized with a risk model of the unknown~\cite{rudd2017extreme}.
In this paper, we introduce an evaluation protocol and baseline agent for this agent-centric approach to HWR.     

\textbf{Related Work.}
Work related to HWR with novelty can be found in the fields of machine learning and computer vision.
There is a strong foundation in deep learning-based approaches to HWR, which have yielded good performance in closed world data set evaluations.
State-of-the-art approaches for diverse document sets~\cite{wigington2017data,german_comp} are based on the Convolutional Recurrent Neural Network (CRNN)~\cite{ShiBY17} in combination with a Connectionist Temporal Classification (CTC) loss~\cite{graves2006connectionist}.

Beyond anomaly detection~\cite{ruff2020unifying}, machine learning work on classifiers has started to look at other ways in which novelty can be handled.
Promising work in this direction relies on statistical modeling using extreme value theory, which more accurately accounts for the samples in the tails of distributions, which is consequential for decision boundaries in classifiers~\cite{bendale2016towards,zhang2016sparse,rudd2017extreme}.
HWR in human biometrics is a mature area of research, having demonstrated that reliable person-specific features exist and can be learned for different languages~\cite{lorigo2006offline,yampolskiy2008behavioural}.
It is an open question as to how well such features work for the characterization of novelty.  

Similar to HWR at large, the problem of writer identification is known to be an unsolved problem, especially in the context of historical documents~\cite{fiel_icdar2017_2017}.
Some works have tried to use ImageNet pre-trained deep neural networks for writer identification as well as other HWR tasks~\cite{studer_comprehensive_2019} and have found improved performance in doing so.
These approaches are built specifically for their task in mind, and tend to not focus on the sharing of information between other HWR tasks as we do in this paper.

\textbf{Contributions.} There are four primary contributions this work makes towards introducing a new challenge problem.
(1) The formalization of HWR with novelty to standardize the domain.
(2) A baseline agent integrating a deep learning-based transcription network and the Extreme Value Machine (EVM)~\cite{rudd2017extreme} for novelty detection.
(3) An evaluation protocol with benchmark data for this domain, including a fully implemented distributed software package for large-scale evaluations\footnote{The code for this paper will be made publicly available after publication at \url{https://github.com/prijatelj/handwriting_recognition_with_novelty}}.
(4) A comprehensive set of experiments including results from over 55,000 experimental tests, setting the state-of-the-art for this challenge problem. 
\section{Formalization of HWR with Novelty}
\label{sec:formal}

The formalization of the HWR domain with novelty includes two parts.
The first is a series of definitions that represent a theory of novelty for the domain.
The second is an ontological specification that characterizes the space of novelty and facilitates measurement of novelty detection difficulty.

\subsection{A Theory of Novelty for HWR}
\label{sec:theory}

The HWR domain, as defined for this paper's proposed benchmark, consists of two high-level tasks: (1) text transcription and (2) style recognition.
The transcription task involves an agent taking a digital image of a handwritten document as input and processing it to recognize the individual characters to produce a plaintext output.
The style recognition task involves the agent identifying known and unknown aspects of visual appearance for both the text (\textit{e.g.}, how are individual characters stylized?) and page (\textit{i.e.}, what does the page look like holistically?).
Two subtasks for style recognition are considered: (2a) writer identification and (2b) overall document appearance identification (ODAI).
The former involves multi-class classification to distinguish between individual known writers and new writers unseen at training time, while the latter involves multi-class classification to distinguish between known global appearances of handwritten documents and appearances unseen at training time.
Any type of novelty can occur in both tasks, thus an important objective of the domain is to detect and manage it.
Note that many other tasks can be defined for transcription and style recognition, and the subsequent theory is general enough to cover those as well.  

A theory of novelty for the HWR domain can be constructed using the recently introduced framework of Boult et al.~\cite{boult2020unifying}, which provides a common basis to define and compare models of novelty across different domains.
In this framework, an agent accesses the world indirectly through a perceptual operator, updating its internal state and acting on the world state as is necessary and possible.
In that regime, the following must be defined: a world, an observational space, perceptual operators, a task, dissimilarity functions to assess potential novelties with respect to the task, and regret functions to determine the impact of incorrect assessments of potential novelties with respect to the task.
A task may consist of multiple subtasks weighted by some priority.

The theory of novelty for HWR extends the image classification theory of novelty, defined in the extended version of the paper by Boult et al.~\cite{boult_unifying_2020_long}. 
As is the case for standard image classification, the samples do not necessarily have any meaningful order.
However, there is a sequential relation of the characters and words within a sample image of a handwritten document.
Below, time step $t$ is in reference to the point in time when a sample image is considered, rather than a character or word within that image.
The following is the specification of the key components that form the theory:

\begin{itemize}
    \item In this paper, an HWR task $\cal T$ can be text transcription, writer identification or ODAI.
    \item A world $\cal W$ in HWR consists of a $d'$-dimensional space of pages of handwritten documents.
    \item An observation space ${\cal O}$ that is accessible to the agent is the $d$-dimensional space that can encode all possible images of handwritten documents.
    This space serves as the agent's feature space, extracted from the image.
    \item The family of perceptual operators ${\cal P}_t$ in HWR are optical sensors, such as cameras, that capture a visible region of the world ${\cal W}_t$, where a time-step $t$ results in a single image of a handwritten document in the case of still image HWR.
    The perceptual operator may continue with feature extraction on the captured image to represent the image as a feature vector $E_t$ of arbitrary dimensions to the agent in the observational space.
    \item The task-dependent world dissimilarity functions ${\cal D}_{w, {\cal T}; E_t}$ and associated novelty threshold $\delta_w$ are determined by ground truth labels associated with the images from the sampling process, where the complete datasets used serve as an oracle (See Supp.~Mat.~Sec.~2.1.1\footnote{\hypertarget{foot:supp_mat}{
        The Supplemental Material is publicly available at 
        \url{https://arxiv.org/abs/2105.06582}
    }}).
    The measurement of dissimilarity uses a distance measure (\textit{e.g.}, Euclidean distance) with a threshold determined by the probability distribution of the data.
    \item The task-dependent observational space dissimilarity functions ${\cal D}_{o,{\cal T}; E_t}$ and novelty threshold $\delta_o$ are determined by the agent's knowledge and design for the task. 
    In a learning-based agent, this is typically done via generalizing from the ground truth in any available training or validation data.
    These functions could make use of Euclidean distance or whichever distance measure suits the observational space.
    \item The world regret function ${\cal R}_{w,{\cal T}}$ is based on the error as measured in the world space given ground truth labels.
    For transcription, this could be Levenshtein Edit Distance, Character Error Rate, or Word Error Rate.
    For the nominal tasks of writer identification and ODAI, the regret is captured by the confusion matrix from which measures of regret derive, such as the normalized mutual information (NMI).
    In this work, NMI is the focus due to ease of interpretability where values near zero indicate poor performance and thus high regret, and values approaching one indicate perfect predictions.
    Of course, any measure, such as NMI, whose value is inversely correlated to regret may have their inverse taken to form a proper measure of regret to be minimized.
    NMI also has a strong information theoretic backing and may be theoretically compared to the NMI of random variables with differing sample spaces, such as continuous random variables.
    \item The observational space regret function ${\cal R}_{o,{\cal T}}$ defines what the agent deems important to the task.
    This is embodied by the agent's internal model for the task, such as the loss function of a neural network or the likelihood calculation in a probabilistic model.
\end{itemize}

Given the above specification, novelty in HWR is deemed to occur in the world when the world ${\cal D}_{w, {\cal T}; E_t}$ exceeds the novelty threshold $\delta_w$.
$E_t$ serves as the history of experience of the agent and plays a key role when a change in the world state is considered novel at a time step.
This paper's baseline agent's $E_t$ is simply the training set and indirect information from the validation set, which informs the agent about how it should set its internal threshold $\delta_o$. 
A world novelty may or may not effect the agent's performance on the task and the novelty's effect may vary in impact to task performance. This is reflected in the world regret ${\cal R}_{w,{\cal T}}$.
The world novelty in any domain, including HWR, must be properly defined in an experiment to assess how an agent performs in its presence.

\begin{figure}[t]
   \centering
   \includegraphics[width=0.7\textwidth]{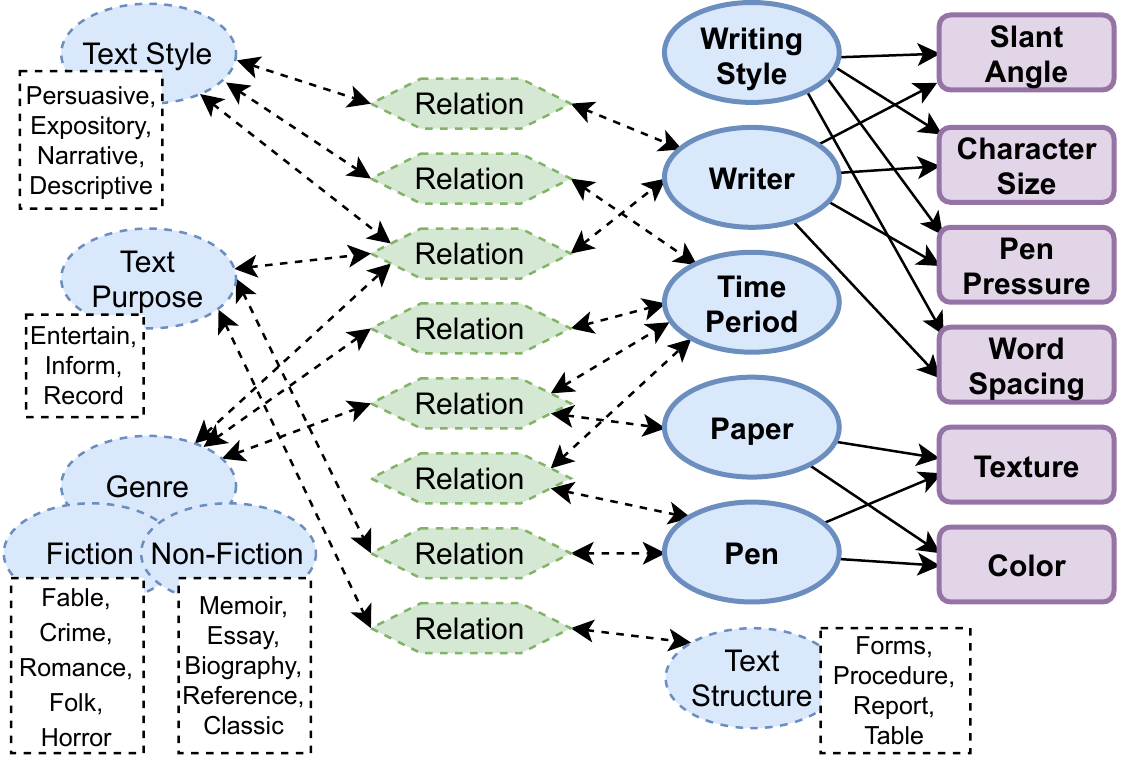}
   \caption{
        HWR Novelty Ontology adapted from~\cite{ontario2006guide}.
        Entities are represented as blue ellipses.
        Entity attributes are represented as purple boxes.
        Examples of entities are dashed, white boxes overlapping the entity node.
        Nodes and edges with solid lines and bold text are the focus of this study.
    }
   \label{fig:ontology_overall}
\end{figure}

In transcription, the sampled world state includes the image and the oracle knows the correct transcription.
World novelty thus occurs in the transcription task whenever a new character, word, or phrase appears in the sequence never before seen in the sampled world at that time-step, given the current label set for each.
For writer identification, world novelty occurs within the handwritten text's style.
In ODAI, world novelty includes a change in document material, backgrounds, perceptual sensor changes, or similar unseen changes in appearance of the document.
World novelty is known in this paper's experiments by the data's labels and dissimilarity is calculated using the confusion matrix or its related measures NMI and accuracy.
The term novelty in this paper is typically in reference to world novelty.

World Novelty is actual novelty that exists in an environment.
For the HWR domain, World Novelty (\textit{e.g.}, novel characters, novel writers, novel backgrounds) is the novelty an agent should be most focused on detecting and managing, because it is actual novelty in the world.
An Observation Novelty occurs when the observation from the perceptual operators of the agent is sufficiently dissimilar from every past observation in the agent's stored experience.
These novelties can only be detected in the HWR domain if the camera acquiring the image of a document has sensed the novelty that is present in the world.
Finally, Agent Novelty occurs when an agent's internal processing cannot map an image to a known state.
In HWR, as described in this paper, Agent Novelty is equivalent to Perceptual Novelty, because we treat state as the predicted label from the perceptual operators. 

Secondary types of novelty exist that are combinations of the above three types.
Unanimous Novelty is the presence of World Novelty, Observation Novelty, and Agent Novelty, and represents a valid transcription or style novelty.
Imperceptible Novelty is novelty that cannot be sensed by the perceptual operators.
In HWR, this can be novelty in the microscopic composition of the material that forms a document page, the historical context in which a document was discovered, the provenance of the document, or any other novelty a camera cannot capture, but a human examiner can determine via other means. 
Faux Novelty is a false positive determination of novelty.
False positives can occur if the perceptual operators encounter noise at acquisition time, injecting novelty into the resulting image that does not exist in the environment.
Further, in perceptual operators for HWR, imperfect machine learning models for transcription and style recognition have error rates, which can also create a false positive situation. 

Regret factors into two additional types of novelty.
Managed Novelty is any novelty that has a minimal impact on agent performance.
In HWR, this could be a change in language expressed by a known character set, which does not impact character transcription performance in any meaningful way.
Nuisance Novelty is a novelty whose world regret and observational space regret significantly differ.
In a document, this could be stain, tear, or other physical artifact on a page that an agent consistently mistakes for a character (thus negatively impacting its error rate), but has little bearing on the environment of the page from the perspective of the world. 

Given this formalization, experiments of HWR with novelty may be designed with a consistent understanding of novelty and its variations.
This provides algorithm designers with an outline for implementing an algorithm given these theoretic constructs.
The formalization establishes a common language to understand how different HWR tasks and their proposed solutions relate to each other.
This allows for ablation studies of agents and understanding how different sources of novelty in HWR images affect performance, as done in the large-scale experiment described in \hyperlink{foot:supp_mat}{Supp.~Mat.~Sec.~5.1}. 

\subsection{Ontological Specification of HWR with Novelty}

An ontological specification serves to describe the knowledge of the world held by the oracle and agent. 
Functionally, the ontology provides terms and structures to reason about and characterize actual and perceived novelty.
We interpret the differences between an agent's task-dependent knowledge of the world and a newly experienced change in the world as a measurable dissimilarity between the world knowledge of the agent and the oracle. 
The degree of dissimilarity forms a basis for assessing the difficulty an agent has in both detecting novelty and performing its task within that novelty space, which is reflected in the expected world regret ${\cal R}_{w,{\cal T}}$.
For example, in the HWR domain, writing samples from a novel writer with a similar style to a known writer are both difficult to detect as being novel and to identify as being written by an unknown writer.

The ontology's components consist of entities, attributes, actions, relations, interactions (passive) and rules often associated with a specific context or domain.
An agent that can detect novelty maintains knowledge elements of the world as described by the ontology.
In closed world supervised learning systems, these knowledge elements are provided through meta-data in the training sets.

The HWR ontology focuses on those components where novelty occurs.
We characterize novelty in terms of text elements including writing style and pen selection, as well as in terms of background elements, \textit{i.e.,} those novelties not specific to the text.
Writing style corresponds to the writer while the last two correspond to ODAI.
The intent of ontological specification is to describe all observable features that may contain novelty including environmental novelties (\textit{e.g.} water damage to the writing medium), temporal and locale novelties (\textit{e.g.} date and time representations and document structures), and text-related novelties, such as copyedit marks.  

The foundational ontology for transcription and style recognition is shown in Fig.~\ref{fig:ontology_overall} (adapted from~\cite{ontario2006guide}).
We focus on a small group of core attributes representative of each ontological entity that best characterizes the set of novelty in the experiments.
We excluded latent attributes from the ontology since they are difficult to qualify in the ontology and beyond the scope of the current study.
However, they may play a critical role in novelty detection and characterization.

The HWR ontology defines these entities and attributes:
\begin{itemize}
\item Four attributes of writer style: (1) pen pressure, (2) slant angle, (3) character size, and (4) word space.  
\item The writer with associations to each style attribute.
\item The image of a handwriting sample.
\item Writing medium (background) categories including types of background noise, textures, and colors.
\item Pen categories including textures and colors.
\end{itemize}

Each component specified by the ontology is associated with a measurement function ${\cal F}_{{\cal O},c}$ applied to each writing sample in the observation space, where $c$ is a category of novelty.
For example, the measurement function for pen pressure is the mean pixel intensity of the written text: ${\left(\sum_{i=1}^{\cal N} \text{pixel}[i]\right)/{\cal N}}$ where ${\cal N}$ is the number pixels in the written text and ${\text{pixel}[i]}$ is the intensity of a pixel ${i}$ in the written text.
The complete set of measures can be found in Table 1 of the \hyperlink{foot:supp_mat}{Supp.~Mat.}
The collection of normalized measures of each component composes a feature vector for use in distance functions (\textit{e.g.}, cosine similarity) to measure the similarity of a novel writing sample to the body of known non-novel writing samples. 

We can also represent the writing style attributes graphically by first creating discrete attributes through binning the component measures and assigning each style and bin to a node in the world knowledge graph (see \hyperlink{foot:supp_mat}{Supp.~Mat.~Sec.~2.2}).
Similarity of writing styles is represented within the knowledge graph by the shared style relations.
The graph supports the application of graph metrics such as isomorphism between two writing style sub-graphs, where a higher number of shared discrete attributes between writing styles indicates a higher similarity.
We hypothesize that the degree of dissimilarity inversely impacts the ability of an agent to detect and characterize novelty.
However, the choice of entities, fidelity for measurement for each ontological entity and weight of significance for each entity impact the utility of dissimilarity measures (see discussion on writer similarity in novel writer discovery in \hyperlink{foot:supp_mat}{Supp.~Mat.~Sec.~5.2}).
\section{Baseline Open World HWR Agent}
\label{sec:baseline_agent}
To show that HWR tasks are feasible in the presence of novelty and to indicate the room for improvement, a baseline open world HWR agent was designed and evaluated on the transcription task and two style recognition subtasks: writer identification and ODAI.
The agent consists of a CRNN~\cite{ShiBY17}, specifically for the transcription task, and an EVM~\cite{rudd2017extreme} for each style subtask.

\begin{figure*}[t]
    \centering
    \includegraphics[width=1.0\textwidth]{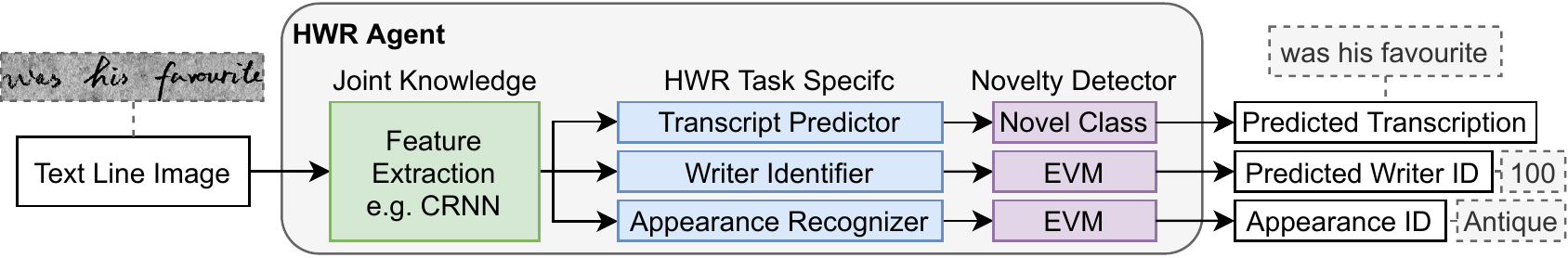}
    \caption{
        The baseline open world HWR agent that is able to detect novelty consists of modules that leverage joint and task-specific knowledge of the HWR domain.
        Thus, it performs the tasks as desired while managing novelty.
        The ``Appearance Recognizer'' is the module for identifying the global appearance of the document, \textit{e.g.}, white background with black foreground, noisy, or antique.
        The ``Novel Class" novelty detector for transcription indicates that the model is trained with a class that represents novelty to isolate those instances from known classes.
    }
    \label{fig:agent}
\end{figure*}

For transcription, the CRNN is trained in a supervised learning fashion on given cropped text lines from documents with the transcribed text as the labels.
The CRNN's architecture consists of a sequence of CNNs leading to a bidirectional LSTM (details are in Supp.~Mat.~Sec.~3.1).
The CRNN covers the feature extraction and transcript predictor modules of the  agent as shown in Fig.~\ref{fig:agent}. Other feature extractors can be used in place of, or in addition to, the CRNN features. 
The CRNN was trained with a single novel character class to manage novel characters and glyphs in the images.

The writer identification and ODAI style tasks are handled by separate EVMs whose supervised training consists of the input of the feature space of a text line image and whose output consists of the labels respective to their task.
For writer identification, the labels are the writer identifiers, and for ODAI the labels are the nominal labels of the overall document image appearance, which is further detailed in Sec.~\ref{sec:experiments}.
The EVMs are classifiers trained to manage novelty in classification tasks and serve as the HWR style task modules and novelty detector modules in Fig.~\ref{fig:agent}.
The EVM predicts a probability vector of the known classes along with the probability of novel class association, given training data.
To further determine if the class is novel, a threshold is applied to the maximum known labels' probabilities.
When the max known class probability is below this threshold, the sample is deemed novel.
The threshold is set by finding the minimum probability that is closest to an equal error rate between the Type I and Type II errors on the training and validation data.

Multiple feature spaces were examined for the style tasks using different feature extraction methods.
These included:
    (1) the mean of the Histogram of Oriented Gradients (HOG)~\cite{dalal2005histograms}, referred to as Mean HOG;
    (2) multiple, evenly spaced, sequential means of HOG over contiguous sections of the image, which is referred to as $M$-Mean HOG where $M$ is the number of sequential means;
    (3) the penultimate layer of ResNet50~\cite{he2016deep} pre-trained on ImageNet~\cite{russakovsky2015imagenet}; 
    (4) the penultimate layer of the CRNN's RNN trained on the transcription task, with PCA applied to reduce dimensionality. We apply this only to the writer identification subtask, as the training of the CRNN is domain specific. 
    
For HOG feature extraction, the resulting values were flattened into a single feature vector.
For $M$-Mean HOG, $M$=10 sequential means were obtained and the Mean HOG vector was appended to the beginning of the 10-Mean HOG feature vector.
These feature extractions were compared to determine which performed best as the image feature representations for the EVMs handling the style subtasks.
See Supp.~Mat.~Sec.~3.1 for more details on this.
\section{Evaluation Protocol and Experiments}
\label{sec:experiments}

For each task described in Sec.~\ref{sec:formal}, a basic experiment was designed to establish baseline performance using the agent described in Sec.~\ref{sec:baseline_agent}.
Two HWR datasets were used to provide data for training and validation, as well as the base data to be transformed for novelty injection.
These were IAM~\cite{marti2002iam} and RIMES~\cite{augustin2006rimes}. 
For each task, IAM and RIMES were split into training, validation, and testing sets.
The data was used in 5-fold cross validation fashion to obtain error bars of the measures.

For 5-fold cross validation, IAM was split by first randomly partitioning the writers into two groups with no intersection of writers, irrespective of sample count.
One half is then split into 5 folds, randomly stratified by the writers to ensure a proper balance of samples representative of those writers' sample frequency.
The remaining half was split into 5 folds of separate writers with no intersection of writers across the folds.
The two halves' folds are paired up, forming the 5-fold split for IAM.
RIMES was simply split randomly into 5 folds because RIMES has no writer identifiers.
This process ensured that every cross validation experiment both shared a subset of IAM writers across all splits, as well as guaranteeing that the validation and testing splits have unique writers unseen during training.
See \hyperlink{foot:supp_mat}{Supp.~Mat.~Sec.~4} for details.

In addition and separate to the above data setup, a large-scale evaluation of the baseline open world HWR agent consisting of approximately 55K tests was conducted using synthetic modifications to the IAM dataset.
For further details on the synthetic modifications, see \hyperlink{foot:supp_mat}{Supp.~Mat.~Sec.~5}.

\textbf{Novel Characters in Text Transcription.}
To evaluate the agent's performance on text transcription, the baseline agent's CRNN was trained on the given text line images with the transcription text as the labels.
The CRNN handled novelty through the addition of a novel character class.
The novel characters in RIMES served as the known unknowns (\textit{i.e.}, labeled negative examples~\cite{scheirer2014probability}) for the CRNN to partially learn the novel character class.
This included characters common in French that are not common in English (\textit{e.g.}, vowels with diacritics).
See \hyperlink{foot:supp_mat}{Supp.~Mat.~Sec.~4.1} for details on the distribution of characters across the data splits.

Character accuracy and word accuracy were used to assess the performance of the agent's transcription.
To assess the novelty detection of the transcription predictor, the presence of novelty was assessed per sample line image.
If a character existed in the ground truth transcription that was not known to the agent, \textit{e.g.,} scratched out characters or vowels with diacritics, then that image was determined to contain novelty.
If the predicted transcript contained a novel character than the line image was predicted to contain novelty.
With this, novelty detection was able to be assessed by a binary confusion matrix from which NMI and accuracy were calculated.

\begin{table*}[t]
    \centering
    \scalebox{0.77}{{\setlength{\tabcolsep}{0.4em}
        \begin{tabular}{l|ll|ll} 
        \textbf{Task} & \multicolumn{2}{c}{\textbf{Multi-class Classif. w/ Novel Class}} & \multicolumn{2}{|c}{\textbf{Binary Novelty Detection}} \\ 
        \multicolumn{1}{r|}{\textbf{Model}} & \textbf{NMI} & \textbf{Accuracy} & \textbf{NMI} & \textbf{Accuracy} \\ 
        \hline
        \textbf{Writer ID} & &  \\
        \multicolumn{1}{r|}{Mean HOG EVM}        & \textbf{0.6462 $\pm$ 1.02e-2} & 0.6524 $\pm$ 6.35e-3    & \textbf{0.6652 $\pm$ 1.22e-2} & \textbf{0.8748 $\pm$ 6.28e-3} \\
        \multicolumn{1}{r|}{10-Mean HOG EVM}     & 0.4127 $\pm$ 7.54e-3 & 0.6932 $\pm$ 4.99e-3     & 0.2199 $\pm$ 5.02e-3 & 0.4425 $\pm$ 6.53e-3 \\
        \multicolumn{1}{r|}{ResNet50 EVM}        & 0.3853 $\pm$ 9.82e-3 & 0.6940 $\pm$ 4.87e-3     & 0.2126 $\pm$ 6.39e-3 & 0.4233 $\pm$ 8.12e-3 \\
        \multicolumn{1}{r|}{CRNN-PCA EVM}        & 0.4013 $\pm$ 1.17e-3 & \textbf{0.7058 $\pm$ 2.73e-3}     & 0.2235 $\pm$ 7.42e-3 & 0.4379 $\pm$ 8.91e-3 \\
        \hline
        \textbf{Appearances (ODAI)} &  & \\
        \multicolumn{1}{r|}{Mean HOG EVM}        & \textbf{0.5713 $\pm$ 3.63e-3} & \textbf{0.7865 $\pm$ 3.59e-3}     & \textbf{0.3262 $\pm$ 7.16e-3} & \textbf{0.6392 $\pm$ 6.48e-3}  \\
        \multicolumn{1}{r|}{10-Mean HOG EVM}     & 0.5065 $\pm$ 5.10e-3 & 0.7383 $\pm$ 8.70e-3     & 0.2955 $\pm$ 5.13e-3 & 0.5747 $\pm$ 6.46e-3 \\
        \multicolumn{1}{r|}{ResNet50 EVM}        & 0.0140 $\pm$ 7.45e-4 & 0.4057 $\pm$ 3.01e-4     & 0.0085 $\pm$ 4.43e-3 & 0.0628 $\pm$ 1.71e-3 \\
        \multicolumn{1}{r|}{CRNN-PCA EVM}        & 0.0551 $\pm$ 1.87e-3 & 0.4092 $\pm$ 5.65e-3     & 0.0345 $\pm$ 2.62e-2 & 0.4354 $\pm$ 4.26e-3 \\
        \hline
        & \textbf{Character Acc.} & \textbf{Word Acc.} & \textbf{NMI} & \textbf{Accuracy} \\ 
        \hline
        \textbf{Transcription} & &  \\
        \multicolumn{1}{r|}{CRNN} & 0.9494 $\pm$ 4.81e-3 & 0.8696 $\pm$ 2.85e-3 & 0.8777 $\pm$ 5.48e-3 & 0.9660 $\pm$ 1.86e-3 \\
        \multicolumn{5}{}{} \\
    \end{tabular}
    }}
    \caption{
    The mean 5-fold results with standard error for the test split of all three experiments.
    ``NMI" stands for Normalized Mutual Information where random guess is 0 and perfect correlation is 1.
    Results for the three tasks indicate that solving the tasks is feasible, but there is still substantial room for improvement.
    Perhaps surprisingly, the Mean HOG dominates across the board for style tasks, except for accuracy for writer identification, but NMI is a more reliable measure of correlation between the labels and predictions.
    All measures reported here are found after selecting the maximum probable class as predicted by the classifier after thresholding the maximum probability to determine if an input is novel.
    }
    \label{tab:mean_res_test}
\end{table*}

The results in Table~\ref{tab:mean_res_test} indicate that the CRNN performs well at text transcription in an evaluation regime that includes novelty. 
The most notable results reside in the novelty detection performance.
The inclusion of the novel character class in the  model for the training data facilitates high scores for NMI and accuracy for novelty detection. However, there is still room for improvement, as these are baseline results.  

\textbf{Novelty in Writer Identification.}
For the writer identification subtask of the style recognition task, all RIMES documents were treated to have an unknown writer because the dataset contains no writer identifiers.
Using the aforementioned data splits, novel writers from the IAM dataset exist across all data splits.
See \hyperlink{foot:supp_mat}{Supp.~Mat.~Sec.~4} for more details on the data breakdown for writer identification.
Based on NMI, Table~\ref{tab:mean_res_test} shows that Mean HOG is the best feature extractor for use with the EVM for writer identification, followed by 10-Mean HOG.
10-Mean HOG includes the same features as Mean HOG in the beginning of its feature vector, so this indicates that the extra 10 sequential means were detrimental to the EVM for writer identification due to too many input variables.
Following closely in third are the features extracted from the CRNN's penultimate layer after principal component analysis (PCA) of 1000 components.
PCA was necessary given memory constraints and due to this a lot of useful information was probably lost, affecting the performance of the MEVM.
See \hyperlink{foot:supp_mat}{Supp.~Mat.~Sec.~3.1} for details.
This is a somewhat surprising result, in that handcrafted features exceed the performance of deep learning in this case.
We analyze this outcome in more detail below.
Overall, the performance for writer identification in the face of novel writers is fair, with rather consistent NMI scores when used as a novelty detector as well.

\textbf{Novelty in Overall Image Appearance.}
Given the above data splits, ODAI involved transforming a subset of the data into different document appearances.
All data splits included the appearances of clean documents, Gaussian noise, antique background, horizontal reflection, and Gaussian blur.
Horizontal reflection and Gaussian blur served as known unknowns to the agent in the training set to give the model some notion of novelty.
To include novelties in the evaluation splits,  vertical reflection was added to both validation and test data splits, and inverted color was added only to the test split.
The differing writers and text in the images introduce nuisance novelty (irrelevant novelty that need not be detected~\cite{boult2020unifying}) into this subtask.
See \hyperlink{foot:supp_mat}{Supp.~Mat.~Sec.~4} for more details on the data split.
ODAI was only performed on the IAM and RIMES datasets, not the synthetically modified IAM dataset.
The agent used an EVM similar to the writer identification task and was assessed using the different feature extractors.
Mean HOG features were again the best. 

\textbf{Feature Extraction Assessment for Style Recognition.}
Mean HOG for feature extraction for the EVM outperformed all other feature extraction methods in Table~\ref{tab:mean_res_test}.
And the second best feature extractor was the 10-mean HOG approach.
However, one might expect that the CRNN, which was trained on the HWR-specific data, would be able to effectively transfer information from the transcription task to the style tasks.
This transfer information may have been limited by the use of PCA, but PCA was necessary due constraints on resources.
Furthermore, the style information is mostly noise for the transcription task, but it is a common source of noise that HWR agents need to manage to generalize onto novel styles.
The CRNN's penultimate layer was probably beyond the point of a useful encoding of style information to being more biased towards the transcription task.

We do see evidence that the CRNN's features provided transfer information for style tasks in its performance relative to ResNet50.
The pretrained ResNet50's penultimate layer as a feature extractor performed the worst out of the batch and this is probably due to limited sharing of information between the ImageNet classification task and the HWR style recognition task.
We suggest that exploiting domain-specific information in a joint manner is still the best strategy in the long run for HWR with novelty.
The key to improving performance is likely a hyperparameter search over many model configurations.
That said, one can still achieve good performance using a combination of handcrafted features and deep learning, with far less computational effort. 

\begin{table}[t]
    \centering
    \scalebox{1.0}{{\setlength{\tabcolsep}{0.4em}
        \begin{tabular}{l|l|ll} 
        \textbf{Task} & \textbf{Novelty Present} & \textbf{NMI} & \textbf{Accuracy}   \\
        \hline
        \textbf{Writer ID} & &  &    \\
        \hline
        \multicolumn{1}{r|}{Mean HOG} & True & 0.1298 $\pm$ 0.0513 & 0.2199 $\pm$ 0.4141    \\
        \multicolumn{1}{r|}{} & False & 0.7768 $\pm$ 0.0345 & 0.7268 $\pm$ 0.4456  \\
        \hline
        \textbf{Transcription} & &
            \multicolumn{1}{l}{\textbf{Character}} 
             & \multicolumn{1}{l}{\textbf{Word}} \\
        && \multicolumn{1}{l}{\textbf{Accuracy}} & 
            \multicolumn{1}{l}{\textbf{Accuracy}} \\      
        \hline
        \multicolumn{1}{r|}{CRNN} & True & \multicolumn{1}{r}{0.6230 $\pm$ 0.2034} &
        \multicolumn{1}{r}{0.5260 $\pm$ 0.2580}
        \\
        \multicolumn{1}{r|}{} & False & 
        \multicolumn{1}{r}{0.8267 $\pm$ 0.1073} &
        \multicolumn{1}{r}{0.7305 $\pm$ 0.2258}\\
        \hline
        \multicolumn{1}{l|}{\textbf{Novelty Detection}} && \textbf{NMI} & \textbf{Accuracy} \\
        \hline
        CRNN \& Mean HOG & True & \multicolumn{1}{r}{0.1300 $\pm$ 0.0999} &
        \multicolumn{1}{r}{0.8029 $\pm$ 0.0981} \\
    \end{tabular}
    }}
    \caption{
    Results for the large-scale 55K evaluation.
    All measures reported here are found after selecting the maximum probable class as predicted by the classifier.
    Novelty Detection is the detection of any novelty in either transcription or writer identification.
    The stark difference between the NMI and accuracy of Novelty Detection indicates the well-known fact that NMI is a better summary measure than accuracy for correlation between the labels and predictions.
    }
    \label{tab:synth_res}
\end{table}

\textbf{Large-Scale 55K Evaluation.}
The large-scale experiment on synthetically modified IAM data adopted a formulation of novelty detection that provides $K$+1-way writer identification.
The agent was trained on a closed set of $K$ writers, identifying newly encountered writers with a single additional class.
Besides writer identification, the agent provides a novelty prediction for each line of text, indicating if the text was produced from  a novel or non-novel distribution of data.

The evaluation was composed of a series of approximately 55K tests.
Each test simulates real world conditions where limited types of novelty are encountered at a specific rate and proportion after a period of no novelty. 
The specification of experimental design, including six independent variables, is described in Table 11 of \hyperlink{foot:supp_mat}{Supp.~Mat.~Sec.~5.1}.

A test is a stream of lines of text sampled from an unrevealed distribution.
In the pre-novelty phase of the test, the agent is presented with small batches of lines written by a subset of $K$ writers.
In the post-novelty phase of the test, the agent is presented with small batches of lines from a new distribution over novel and non-novel writing samples. To eliminate early false positive novelty indications, the agent establishes a prior distribution composed of non-novelty lines presented in a pre-novelty phase of the test.
The agent weights instance-level novelty predictions based on a distribution shift of presented lines, signaling entry into a post-novelty phase. 

In each test, novel writing examples are sampled from collections containing one type of novelty --- either unknown writers, novel pens or novel backgrounds.
The novel elements of the writing samples exemplify open world variations in choice of writing instrument and medium.
A high-level summary of the results can be found in Table.~\ref{tab:synth_res}, and a more detailed analysis in \hyperlink{foot:supp_mat}{Supp.~Mat.~Sec.~5}. 
Results indicate that solving the tasks is feasible, but given the much larger scale of evaluation compared to the experiments in Table~\ref{tab:mean_res_test}, there is much room for improvement. This leaves an opening for new work on this problem.
\section{Conclusion}
This paper introduced an agent-centric approach to handling novelty in the HWR domain. This domain is attractive for the study of novelty, as it consists of a key challenge problem within AI: reading in a more human-like way.
The HWR domain with novelty was formalized, an evaluation protocol with benchmark data was introduced, and comprehensive results from a baseline agent were presented to provide the research community with a starting point to build upon.
Beyond incremental improvements in transcription performance and style recognition in the presence of novelty, we suggest that adaptation via incremental learning is the next step. Agents that can properly react to and manage novelty, as opposed to merely detecting novelty, will perform better on the task over time. With additions to the evaluation protocol supporting this, we expect a new class of agents to appear for a number of document processing applications.

\bibliographystyle{splncs04}
\bibliography{bibs/background, bibs/data, bibs/theory}

\end{document}


\title{
    \papertitle: Supplemental Material\thanks{This research was sponsored by the Defense Advanced Research Projects Agency (DARPA) and the Army Research Office (ARO) under multiple contracts/agreements including HR001120C0055, W911NF-20-2-0005,W911NF-20-2- 0004,HQ0034-19-D-0001, W911NF2020009. The views contained in this document are those of the authors and should not be interpreted as representing the official policies, either expressed or implied, of the DARPA or ARO, or the U.S. Government.}
}
\titlerunning{\papertitleshort: Supplemental Material}
\author{
    Derek S. Prijatelj\inst{1}\orcidID{0000-0002-0529-9190}
    \and Samuel Grieggs\inst{1}\orcidID{0000-0002-2433-5257}
    \and Futoshi Yumoto\inst{2}\orcidID{0000-0001-5354-8254}
    \and Eric Robertson\inst{2}\orcidID{0000-0002-4942-4619}
    \and Walter J. Scheirer\inst{1}\orcidID{0000-0001-9649-8074}
}

\authorrunning{D. Prijatelj et al.}

\institute{
        University of Notre Dame, Notre Dame IN 46556, USA\\
        \email{\{dprijate, sgrieggs, walter.scheirer\}@nd.edu}
    \and
        PAR Government, 421 Ridge St, Rome NY 13440, USA\\
        \email{\{eric\_robertson, futoshi\_yumoto\}@partech.com}
}

\maketitle

\section{Supplemental Material}

This is the supplemental material for the main paper intended to provide complete details of the Handwriting Recognition (HWR) domain and agent-centric approach to it for others interested in working on this challenge problem.
Additional notes are provided for the domain formalization  and the evaluation protocol.
A large selection of supplemental experiments is provided to explore different aspects of the HWR domain with novelty in more detail.   

\section{Formalization of HWR with Novelty}

This section consists of further details on the formalization of the HWR domain. Specifically, we provide more discussion on the HWR novelty theory, oracle definition, and ontology here. 

\subsection{A Theory Novelty for HWR: Additional Details}

The theory of novelty for the HWR domain in the main paper notably excludes two pieces from Boult et. al~\cite{boult2020unifying}.
One of those components is the agent's ($\alpha$) action space $\cal A$ that contains all possible actions $a_t \in {\cal A}$ that the agent may take.
The other part is the state recognition function $f_t(x_t, s_t): \R^{d} \times {\cal S} \mapsto {\cal S} \times {\cal A}$, where $x_t$ is an observation-space input, and $s_t \in {\cal S}$ is the agent's state at the current time-step $t$ for all possible states $\cal S$ of the agent.
In traditional image classification tasks, these two components do not come into play as the agent is neither stateful nor does it take actions in the world beyond outputting the predicted class for a sample.
In image classification, as well as the HWR domain with novelty as defined in the main paper, the state recognition function could simply be the predicted class for a given sample.

However, these two components could be a part of an agent for the HWR domain with novelty.
For example, the state of the agent could update if the agent were to include incremental learning to solve one of the HWR tasks.
The incremental learning would then be a state change in the learning for the agent, as would any ``learning" that were to occur over time, such as the continued training of an artificial neural network.
An extreme example of incremental learning would be an agent that is pre-trained on available handwritten documents, but is given a new set of documents in a different language, thus with new glyphs, characters, words, language, etc., to be learned and is only given these new documents over time as they are discovered.

If the open world HWR agent proposed in the main text were to be extended and given the ability to act in the world, then the possible actions $\cal A$ would come into play.
An example of this would be an automated robot performing the entire handwritten document transcription process itself, \textit{e.g.}, opening the physical books carefully, gently turning the pages, and possibly changing the perceptual operator by zooming in and out of the images or changing sensors.

\subsubsection{2.1.1 Caveats of Defining the Oracle}
The oracle mainly consists of the datasets used for evaluation.
However, the oracle also determines world dissimilarity and regret functions given the data.
For datasets that consist of ground truth labels typically used in supervised learning, the datasets may be enough.
However, there may be additional domain knowledge that is not available in the datasets, either implicitly or explicitly defined by the oracle.
An example is the case of label frequency within the dataset mismatching the current domain knowledge of the world.
In this case, the oracle provides label weightings to adjust the dataset's samples to better represent the current domain knowledge of the universal population of those labels.
The information about these weightings may or may not be provided to the agent as an extended part of the world space.

Another example case is when the dataset has missing labels, either partially or completely, the oracle is expected to provide the information that defines the task given the data.
A complete lack of labels for a certain type of information is a rather common occurrence in domains where novelty is present and cannot be labeled in any capacity, and where labeled or unlabeled data may be used in training and evaluation.
The datasets do not necessarily define the oracle's information in its entirety.
This is a challenge for evaluation design that needs to be accounted for when assessing agents that must manage novelty.
A sampling problem, such as HWR, thus defines the oracle and task through the world space, world dissimilarity, world regret, and the information used from datasets and domain knowledge.

    

\subsection{HWR Ontological Specification}
\label{sec:hdt_ontology}

An ontological specification serves to characterize the novelty space and provide a basis for measuring the difficulty of detecting novelty within that space.
Novelty in HWR is organized by three categories: writing style, pen selection and background novelties, which are the novelties that are not specific to the content of the text.
The intent of specification is to describe all observable novelties (from an oracle's view of the world) including environmental novelties such as water damage to the writing medium, temporal and locale novelties such as date and time representations and document structures, and text-related novelties such as copyedit marks. 

\begin{figure}[t]
\centering
  \includegraphics[scale=0.40]{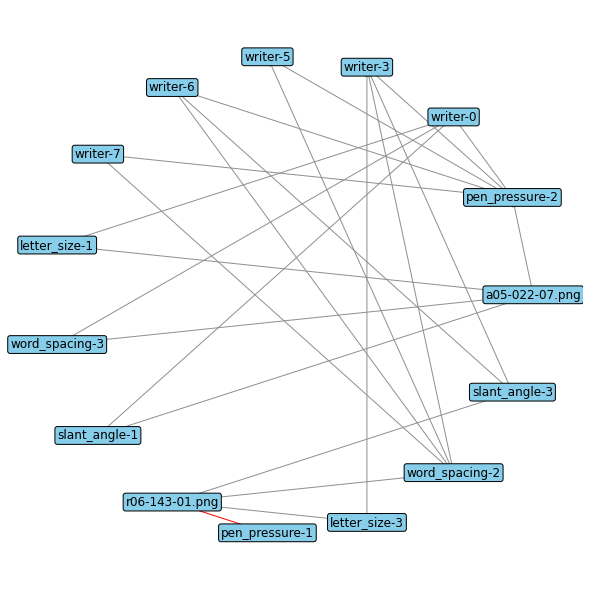}
  \caption{
    Illustrative Knowledge graph of Writing Style for four style attributes associated writing samples a05-022-07 and r06-143-01 and five selected writers.
    The red edge on the bottom represents the writing style of a sample not associated with the sample's author.
  }
  \label{fig:knowledge_style}
\end{figure}

In the development of the ontology-based knowledge graph, we first start with the characterization of writing style.
Writing style is made up of the style attributes slant angle, word space, character size and pen pressure.
Each style attribute is described by a continuous function, defined in Table~\ref{tab:style_measurements}, which is applied to images of words present in each writing sample~\cite{malemm2018}.
The results from the functions for all samples are binned to form discrete style descriptors, which are used to construct style attribute nodes in the knowledge graph.
The number of bins is chosen to provide adequate separation of each writing style.
In our initial assessment, we used four bins for slant angle and three bins for the rest of the style attributes.
The style attributes are collected for all writing sample images.
The most frequent style attribute value is assigned to each writer.
The result is a set of associations between each writing sample, the style, and the writer, as shown in the knowledge graph in Fig.~\ref{fig:knowledge_style}.
We apply the same approach for background and pen novelties.
The non-style measurement functions for these novelties are described in Table~\ref{tab:non_style_measurements}.

\begin{table}
    \centering
    \begin{tabular}{|l | l|} 
        \hline
        \textbf{Style} & \textbf{Function} \\      
        \hline
        Pen Pressure & 
        \vtop{\hbox{\strut ${\left(\sum_{i=1}^{\cal N} pixel[i]\right)/{\cal N}}$}\hbox{where \strut ${\cal N}$  is the number pixels} \hbox{in the written text,} \hbox{and \strut ${pixel[i]}$ is the intensity of a pixel ${i}$.}} \\    
        \hline
        Slant Angle &  \vtop{\hbox{\strut ${\max\limits_{A^i} ~S(A^i)}$} \hbox{where ${A^i}$ is the set of angles} \hbox{[-45,-30,-20,-15,-5,0,5,15,20,30,45],} \hbox{ and ${S(A^{(i)})}$ is a shear estimate} \hbox{~\cite{Vinciarelli2020}.}}   \\ 
        \hline
        Word Spacing & \vtop{\hbox{Average number of horizontal pixels}\hbox{between words where a space is a vertical} \hbox{slice with fewer than 30\% quantile} \hbox{of vertical pixels for a line image.}}  \\   
        \hline
        Character Size & \vtop{\hbox{Average number of pixels over all} \hbox{vertical slices of the image} \hbox{excluding those slices labeled as a space.}}   \\   
        \hline
    \end{tabular}
    \caption{Style Measurement Functions}
    \label{tab:style_measurements}
\end{table}

\begin{table}
\centering
    \begin{tabular}{|l|l|} 
        \hline
        \textbf{Novelty Type} & \textbf{Function} \\      
        \hline
        Background & 
        \vtop{\hbox{Entropy of the grey level} \hbox{background (without text)}} \\    
        \hline
        Pen &  \vtop{\hbox{Entropy of the grey level} \hbox{pixel intensities in the written text.}}   \\ 
        \hline
    \end{tabular}
    \caption{Non-Style Measurement Functions.}
    \label{tab:non_style_measurements}
\end{table}

A correct knowledge graph consists of each writing sample associated to a single writer via a two step path through the four style attribute nodes.
The writing style measurement functions provide a gross measure of writing style.
Combined with the inaccuracies introduced with binning, not all writing samples from the same writer are associated with the same set of bins across all style attributes.
Since the same writer's style is an aggregate value over a set of writing samples, some binned measures for a writing sample form an association to a style attribute not associated with the writer of the sample.
This is highlighted in the sample graph in Fig.~\ref{fig:knowledge_style} via a red edge.
This suggests an optimization strategy for style binning and association to maximize the number of writing samples associated with a writer through the four style attribute nodes.

\subsection {Ontological Specification for Novelty Characterization}

\begin{figure}[t!]
    \centering
  \includegraphics[scale=0.35]{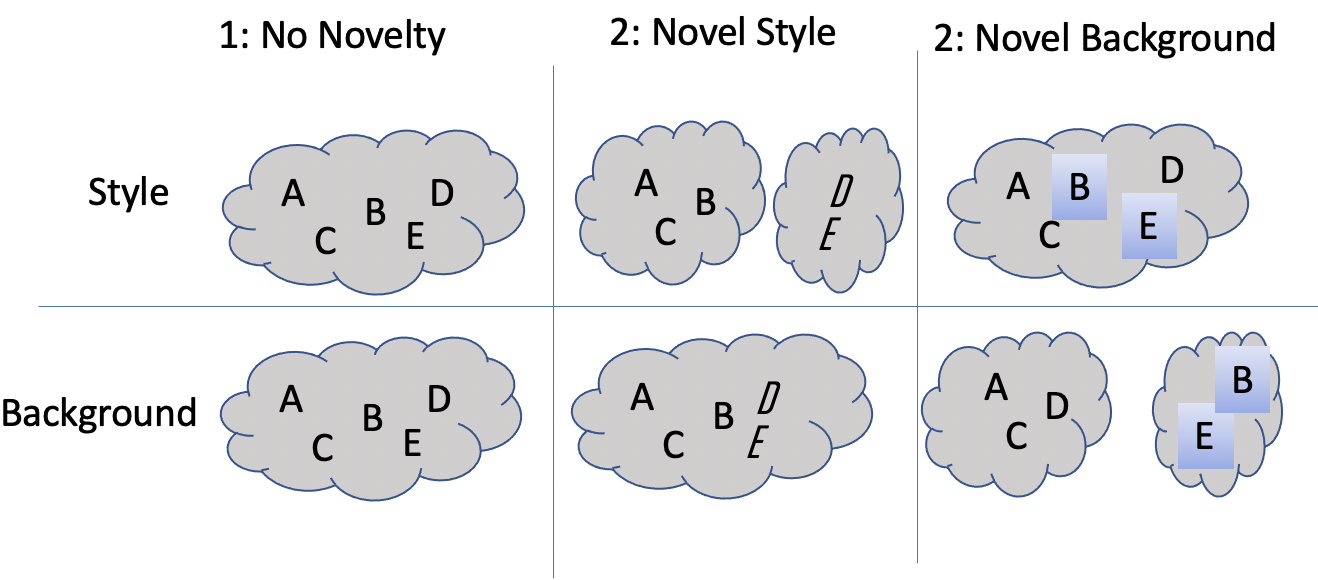}
  \caption{Example Clusters for two cluster groups, Style and Background, under three novelty types: (1) No Novelty; (2) Novel Style; and (3) Novel Background.}
  \label{fig:clusters}
\end{figure}

Characterization was achieved through groups of clusters over writer samples created by the agent.
Each group explains a single characterization of novelty as it occurs in each text image.
Groups included in our initial study are:
\begin{itemize}
  \item Up to 3 clusters for pen pressure, character size and word spacing,
  \item Up to 4 clusters for slant angle,
  \item Up to 3 clusters for category of novelty: writer novelty, background and pen novelties.
\end{itemize}
A single `writer novelty' cluster occurs in the novelty category cluster group when novelty does not occur --- all non-novel examples cluster together.
Fig.~\ref{fig:clusters} illustrates this approach with two cluster groups.

For performance evaluation of characterization, we use Normalized Mutual Information (NMI) to measure the quality of the clusters.
We first separate the agent characterizations of writing samples with no novelty and the three categories of novelty: writing style, pen and background.
We interpret characterizations in the non-novel subgroup as a base measurement of the agent's dependence on the cluster-represented attributes to describe novelty.

\begin{table}[b]
\centering
\begin{tabular}{|l | r r r r r |}
\hline
\textbf{Novelty} & \textbf{PP} & \textbf{CS} & \textbf{WS} & \textbf{SA} & \textbf{NC} \\
\hline
Style        &  ${PP_s}$ & ${CS_s}$  & ${WS_s}$ &${SA_s}$ & ${NC_s}$ \\
\hline
Background   & ${PP_b}$ & ${CS_b}$ & ${WS_b}$ & ${SA_b}$& ${NC_b}$ \\
\hline
Pen          & ${PP_p}$ & ${CS_p}$  & ${WS_p}$ &${SA_p}$ & ${NC_p}$ \\
\hline
No Novelty   & ${PP_n}$ & ${CS_n}$  & ${WS_n}$ &${SA_n}$ & ${NC_n}$ \\
\hline
\end{tabular}
\caption{Characterization cluster groups are Pen Pressure (PP), Character Size (CS), Word Spacing (WS), Slant Angle (SA), and Novelty Category (NC). }
\label{tab:characterization}
\end{table}

The characterization promotes better understanding of an agent's performance in the HWR domain with novelty.
We first establish a baseline cluster quality using non-novel writing samples.
In the baseline, cluster groups organize samples by similar styles and backgrounds.
As different types of novelty are introduced, new cluster centers are formed to isolate those samples perceived as having the group's representative novelty. 

We partition and evaluate the characterization clusters by novelty category, shown in each row of Table~\ref{tab:characterization}, to highlight the interactions between different categories of novelty and the novel style attributes.
Applicable measures to this structure include NMI and cluster purity.
This structure for analysis aids in understanding agent response to mixed novelties, such as style and background changes.
The No Novelty row serves as a baseline characterization of writing samples without novelty.
The Style row measures characterization clusters of samples with novel writing styles.
In terms of a mapping to empirical observations, low performance for cell $PP_p$ (Pen, Pen Pressure) in comparison to baseline No Novelty was observed, indicating that this paper's open world agent is unable to discern pen changes from pen pressure novelty.  We do not expect to see much variation from the non-novel examples in cells ${CS_p}$, ${WS_p}$ and ${SA_p}$,  since pen pressure novelties do not significantly affect character size, slant angle and word spacing. We also expect matched performance to the baseline No Novelty conditions for separable novelty categories, such as all Style cells in the Background row ($PP_b$, $CS_b$, $WS_b$, $SA_b$). For example, an agent is expected to separate novel from non-novel backgrounds, but may fail to adequately separate groups of samples using two different novel backgrounds.   

The Novelty Category (NC) cluster group serves to characterize the core types of novelty.  NC cells $NC_c$, $NC_b$ and $NC_p$ are meant to be measurements of an agent's ability to distinguish different samples within the same category of novelty.  For example, $NC_b$ measures an agent's ability to distinguish examples with blue backgrounds from those with red backgrounds.   

For an initial assessment, we characterized the last 32 test images selected from each test prior to evaluating characterization of the novelty.  We provide the sample set of measurements using Cluster Purity in Table~\ref{tab:characterization_baseline}.

\begin{equation}
 \text{Purity} = \frac 1 N \sum_{i=1}^k max_j | c_i \cap t_j |
\end{equation}
 where ${N}$ = number of writing samples, ${k}$ = number of clusters, ${c_i}$ is a cluster in ${C}$, and ${t_j}$ is a ground truth novelty label.

In this sample, we see evidence of confounding variables when characterizing pen pressure with pen changes. Characterization of slant angle, when compared to non-novel cases, is weakly affected by pen changes. Word spacing was significantly affected in all three novelty cases. Style changes were correctly separated from background and pen novelties, as indicated in the Novelty Category cluster group.

\begin{table}[h]
\centering
\begin{tabular}{|l | r r r r r |}
\hline
\textbf{Novelty} & \textbf{PP} & \textbf{LS} & \textbf{WS} & \textbf{SA} & \textbf{NC} \\
\hline
Style        & 0.88 & 0.85 & 0.55 & 0.53 & 1.00 \\
\hline
Background   & 0.83 & 0.89 & 0.45 & 0.61 & 0.89 \\
\hline
Pen          & 0.75 & 0.71 & 0.57 & 0.77 & 1.00 \\
\hline
Non-Novel    & 0.84 & 0.75 & 0.80 & 0.80 & 1.00 \\
\hline
\end{tabular}
\caption{Cluster Purity Characterization Results based on novelty type.  Characterization cluster groups here are Pen Pressure (PP), Letter Size (LS), Word Spacing (WS), Slant Angle (SA) and Novelty Category (NC).}
\label{tab:characterization_baseline}
\end{table}

\section{Additional Information on HWR Agents}
The details of the baseline open world HWR agent generally defined in the main paper are included here, along with an additional agent that is not designed to handle novelty.

\subsection{Baseline Open World HWR Agent}
This paper's proposed HWR novelty detecting agent is further specified in this section.
This baseline open world agent for HWR is built upon the Convolutional Recurrent Neural Network (CRNN) architecture which is commonly used for closed set HWR tasks.
The IAM dataset~\cite{marti2002iam}, a very commonly used handwritten text dataset, contains a number of writing errors that were introduced when the dataset was created.
These are mostly in the form of crossed out misspelled words.
The ground truth provided with the dataset represents these errors, with the ``\#" character.
This is treated as the baseline agent's exposure to novel characters, serving as the known unknown class.
This known unknown class was further expanded upon by introducing a subset of novel characters from the RIMES dataset, which contains numerous characters with diacritics that are not present in IAM.

For all experiments using a CRNN, the model was structured as follows and is shown in Table~\ref{tab:crnn_model}.
Five convolutional layers feed into five bidirectional LSTM layers, each with a kernel size of 3 and a stride and padding of 1.
The 5 LSTM layers have a hidden size of 256. Input Images were resized to 64 pixels tall.
The CRNNs were trained until they did not improve for 80 epochs using the RMSprop optimizer with an initial learning rate of $3*10^{-4}$.
On average the models would train for around $300$ epochs, using a batch size of $8$. Training proceeded indiscriminately on a selection of Nvidia Titan X, Titan Xp, 1080ti, 2080ti and RTX 6000 GPUs, with each epoch taking about 5-10 minutes depending on the GPU used. Inference averaged about 33 milliseconds per sample on a 2080ti.

The CRNN serves as both the feature extractor and transcript predictor as seen in the main paper's Fig.~2.
In the supplemental Figure~\ref{fig:transcript_crnn}, the CRNN portion of the agent is depicted in isolation to indicate its merging of feature extraction and transcript predicting during its training in a supervised learning fashion.
To be used as a feature extraction for the EVMs for writer identification and ODAI, the penultimate layer of the CRNN (specifically the last layer of its RNN portion), was used as the encoding of the line images.
Given that the sequence differs with the length of the input line image, the encodings were padded with zeros to the maximum time-step size (656 with the IAM and RIMES data) and then an incremental principal component analysis~\cite{ross_incremental_2008} method was used with 1000 components to obtain a memory-managable encoding for the EVMs.
And to expedite the process given both memory and time constraints we used only one fourth of the data after running through the CRNN.
Due to using 1000 components to fit in memory, a lot of useful information was lost and probably significantly affected the performance of the MEVMs that used the CRNN-PCA as their feature space.

\begin{figure}[t]
    \centering
    \includegraphics[width=1.0\textwidth]{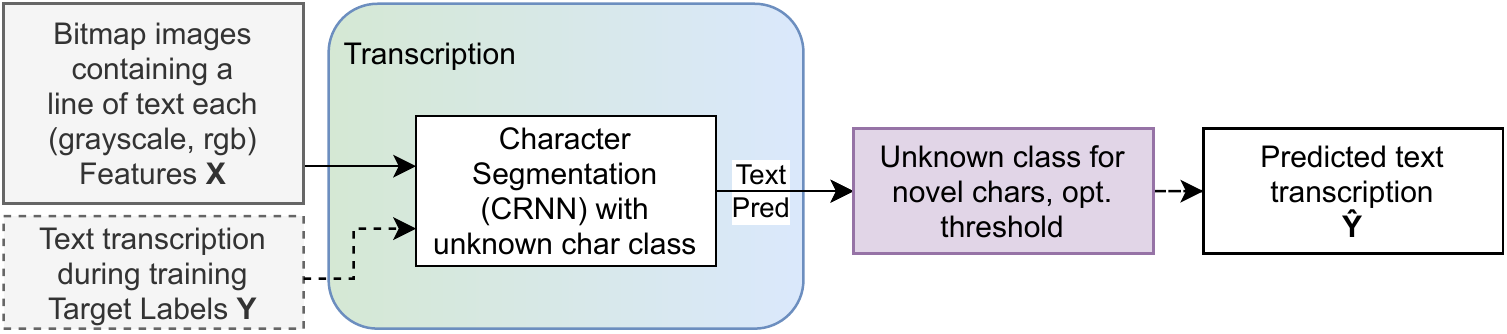}
    \caption{
        The CRNN in isolation depicting its joint use as both the feature extractor and transcript predictor.
        The feature extraction occurs in joint with its supervised learning of the transcription task.
        The penultimate layer of the CRNN is used as one of the examined feature spaces for training the style task EVMs, after using PCA on the zero padded sequential output.
        Due to memory constraints, 1,000 components were used with from an incremental implementation of PCA on 25\% of the training data.
    }
    \label{fig:transcript_crnn}
\end{figure}

The Extreme Value Machine (EVM) is an open set classifier designed to handle novel classes~\cite{rudd2017extreme}.
The EVM has various hyperparameters, including tail size, cover threshold, distance measure, and distance multiplier.
The tail size used was 1000, the cover threshold was 0.5,  the distance measure used was Cosine similarity, and the distance multiplier was 0.5.
These hyperparameters were the same for the two separate EVMs trained on their respective style tasks of writer identification and ODAI.
The implementation of the EVM used is a pytorch version with GPU support\footnote{This will be made publicly available 
after the publication of this paper.}.
EVM training took approximately 2 hours per training fold on both data sets, thus equating to 20 hours total of EVM training given the two experiments with 5-fold cross validation.
Prediction time was approximately one half hour for each evaluation fold using the EVM.
Hardware used for training and inference of the EVM matched that of the CRNN described above.

The EVM's output a probability vector of size $K+1$ for $K$ known labels.
The extra label serves as the general novel class label (referred to as the unknown class in the EVM documentation).
To obtain this probability vector, the EVM outputs all of the probabilities for the $K$ classes in its implementation.
To obtain the probability of the novel class, the probability of the maximum probable known class $k_m$ is taken with $1 - k_m$ calculated as the probability of the novel class.
The rest of the known probabilities are scaled by $k_m$ and the probability of novelty is appended to the end of the probability vector.

\subsection{Baseline Closed World HWR Agents}

Two additional closed world agents were evaluated as comparison points to the open world agent. One agent performs the writer identification style subtask, while the other performs the text transcription task. They do not pass information between each other, and have no specific abilities to manage novelty. 

\subsubsection{Baseline Closed World Writer Identification Agent}

A baseline closed world agent for just writer identification was created for comparison to the open world agent, described in the main text, under novel conditions. The closed world baseline agent predicts, for each sample, one of the 50 known writers in the training set by applying the softmax function to the output of the dense layer of a CNN.  The baseline model serves to demonstrate limited utility only in a closed world, over-fit to known writers, with considerable degradation in performance when exposed to novel conditions.

The baseline writer identification model is a neural network consisting of three groups of 2-D convolution layers with RELU activation and max pooling, followed by two groups dense connected layers with RELU activation and 50\% drop out, ending with a dense softmax activated layer over the 50 known writers~\cite{reddy2018}.

\begin{table}[t]
    \centering
    \begin{tabular}{|l c|} 
        \hline
        zero\_padding2d & (115, 115, 1) \\      
        \hline
        Conv-2D & (58, 58, 32) \\    
        \hline
        MaxPooling2D &  (29, 29, 32) \\ 
        \hline
        Conv-2D & (29, 29, 64)  \\   
        \hline
        MaxPooling2D &  (14, 14, 64) \\   
        \hline
        Conv-2D & (14, 14, 128) \\    
        \hline
        MaxPooling2D &  (7, 7, 128) \\  
        \hline
        Flatten & 6,272 \\ 
        \hline
        DropOut & 6,272 \\ 
        \hline
        Dense & 512 \\ 
        \hline
        DropOut & 512 \\ 
        \hline
        Dense & 256 \\ 
        \hline
        DropOut & 256 \\ 
        \hline
        Dense & 50 \\ 
        \hline
    \end{tabular}
    \caption{Baseline Closed World Writer-Predictor Agent Model.}
    \label{tab:baseline_model}
\end{table}

\begin{table}
    \centering
    \begin{tabular}{|l c|} 
        \hline
        Conv2d & (16, 64, x) \\    
        \hline
        BatchNorm2d & (16, 64, x) \\ 
        \hline
        LeakyReLU & (16, 64, x)\\
        \hline
        MaxPool2d & (16, 32, .5x) \\   
        \hline
        Conv2d & (32, 32, .5x) \\   
        \hline
        BatchNorm2d & (32, 32, .5x) \\    
        \hline
        LeakyReLU & (32, 32, .5x) \\  
        \hline
        MaxPool2d & (32, 16, .25x) \\ 
        \hline
        Conv2d & (48, 16, .25x) \\ 
        \hline
        BatchNorm2d & (48, 16, .25x) \\ 
        \hline
        LeakyReLU & (48, 16, .25x) \\ 
        \hline
        Dropout2d & (48, 16, .25x) \\ 
        \hline
        Conv2d & (48, 16, .25x)\\ 
        \hline
        BatchNorm2d & (48, 16, .25x)\\ 
        \hline
        LeakyReLU & (48, 16, .25x)\\
        \hline
        Dropout2d & (48, 16, .25x)\\
        \hline
        Conv2d & (64, 16, .25x)\\
        \hline
        BatchNorm2d & ((64, 16, .25x)\\
        \hline
        LeakyReLU & (64, 16, .25x)\\
        \hline
        Conv2d & (80, 16, .25x)\\
        \hline
        BatchNorm2d & (80, 16, .25x)\\
        \hline
        LeakyReLU & (80, 16, .25x)\\
        \hline
        Flatten Interior & (1280, .25x)\\
        \hline
        Reshape & (.25x, b, 1280)\\
        \hline
        5 x bidirectional LSTM & (.25x, b, 512)\\
        \hline
        \hline
        Linear & (.25x, b, 80)\\
        \hline
        LogSoftmax & (.25x, b, 80)\\
        \hline
    \end{tabular}
    \caption{
        Convolutional Recurrent Neural Network used for Handwriting Recognition in the Baseline Open World Agent.
        For experiments in which a CRNN embedding is used, the embedding is extracted at the double line.
        Note that `x' is the input image width and `b' is the batch size, which is only shown when it is not in the first position.
    }
    \label{tab:crnn_model}
\end{table}

\begin{table*}[t]
    \centering
    \scalebox{0.95}{{\setlength{\tabcolsep}{0.4em}
        \begin{tabular}{|l|l|lll|lll|} 
        \multicolumn{8}{c}{\textbf{IAM Writer Identification Distribution for}} \\ 
        \multicolumn{8}{c}{\textbf{Basic Feature and Transcription Evaluation}} \\ 
        \hline
         Datasplit & Total
            & \multicolumn{3}{c|}{\textbf{Total Writers}}
            & \multicolumn{3}{c|}{\textbf{Total Intersecting}} \\ 
         Type & Writers  
            & \multicolumn{3}{c|}{\textbf{in Split}}
            & \multicolumn{3}{c|}{\textbf{Between Pairs}} \\ 
         &   & train & val & test & train \& val & train \& test & val \&test  \\ 
         \hline
         IAM Aachen        & 431 & 373 & 93 & 170 & 65 & 135 & 57  \\ 
        \hline
         5-Fold CV   & 431 &  $\sim$354 & $\sim$251 & $\sim$259 &  $\sim$216 & $\sim$216  & $\sim$216 \\ 
        \hline
    \end{tabular}
    }}
    \caption{
        The mean 5-fold cross validation experiment's distribution for IAM writer identification.
        The version of the IAM data is the Aachen version, which standardizes some of the character transcriptions and handles errors.
        RIMES was excluded from this due to all RIMES documents being treated as a single unknown writer.
        The used 5-fold cross validation indicates the approximate distribution of writers between each data split for a single round of training and evaluation of a fold.
    }
    \label{tab:iam_writer_distrib}
\end{table*}

\subsubsection{Baseline Closed World Text Transcription Agent}

A baseline closed world agent for just text transcription was created for comparison to the open world agent described in the main text under novel conditions. This baseline agent produces text for each writing sample, based on what it knows from the 50 known writers in the training set, by applying log-softmax to the output of deep recurrent layers~\cite{shi2015endtoend}.  It serves to demonstrate limited utility only in a closed world, over-fit to known writers, with considerable degradation performance when exposed to novel conditions.

\section{Basic Feature and Transcription Evaluation: Additional Protocol Information and Detailed Analysis}

The data splits for the cross validation were obtained by first halving the unique writers into two equal groups of 216 each.
Then, one half was randomly shuffled and split into 5 folds in a traditional 5-fold cross validation manner, stratified by the writer identifiers for the best representation of all 216 writers in each fold's samples.
The other half was then further split into 5 groups of unique writers with no intersection.
This second split ensures that for every fold, there is a set of novel writers in the test dataset.
Each half's 5 folds were then aligned randomly such that typical 5-fold cross validation may occur.
The training set, consisting of 4 folds, for every round of cross validation was then split in the exact same way, such that the validation set also had novel writers at evaluation time.
This method of obtaining the 5-fold cross validation folds results in the approximate distribution of writers in training, validation, and testing sets as seen in  Table~\ref{tab:iam_writer_distrib}.
This is approximate, because due to the imbalanced number of samples per writer, where some writers had less than 5 samples, some folds had more writers than others.

\subsection{Novel Characters in Transcription}
In general, most HWR models will have some sort of ``background'' class to represent spurious marks or mistakes on the page.
For the purposes of this experiment we trained on a combined IAM and RIMES dataset in which the RIMES transcriptions were modified to include the characters that are not a part of the IAM dataset as background.
RIMES and IAM were broken into folds such that Zipf's law gives us a variety of known unknown and unknown unknown characters in each fold, in the terminology of open world recognition.
In terms of novelty, due to Zipf's law, novel characters unseen at training time occurred naturally in both the validation and testings sets for all folds.
The addition of RIMES, and thus all of its French specific characters not included in the IAM dataset, included more characters  whose labels were never known to the agent.
However, some were included by design in the training set as unknown characters seen during training (known unknowns).

\subsection{Novelty in Overall Image Appearance}

In order to simulate novelty in the ODAI style recognition subtask, we augmented the IAM and RIMES datasets by randomly modifying the backgrounds of the images.
The data was split into three different representation classes for training, Noise, which added Gaussian background noise as well as over the foreground, Antique, which adds a background similar to that of a historical document, and the Original White background from a clean document scan.
The Original White background with black foreground is the unaltered background of each image found in IAM and RIMES, and is the typical ideal clean scan of handwritten documents.
Additionally there are two known unknown classes that are seen at training time, Reflect\_0, which is flipping the text image over the horizontal axis, and Blur, which adds Gaussian blur to the image.
There is another augmentation only included in the validation and test sets, Reflect\_1, which reflects the image over the vertical axis. 
Finally, the test set includes another novel image appearance class where the image has inverted color, the InvertColor class.

The Antique class used a set of free-to-use background images totaling in 16 different background images all accessed as of the data 2020-12-30.
Nine of these images were taken from commons.wikimedia.org that were categorized as ``old paper", ``vintage paper", or as ``parchment":
\begin{itemize}
    \item El siglo de las tinieblas, o memorias de un inquisidor; novela hist\'orica origina:
        \href{https://commons.wikimedia.org/wiki/File:El\_siglo\_de\_las\_tinieblas,\_o\_memorias\_de\_un\_inquisidor;\_novela\_hist\%C3\%B3rica\_original\_(1868)\_(14590934239).jpg}
        {
            https://commons.wikimedia.org/wiki/File:El\_siglo\_de\_las\_tinieblas,
            \\\_o\_memorias\_de\_un\_inquisidor;\_novela\_hist\%C3\%B3rica\_original\_(1868)
            \\\_(14590934239).jpg
        }
    \item Old paper 1:
        \url{https://commons.wikimedia.org/wiki/File:Old\_paper1.jpg}
    \item Old paper 3:
        \url{https://commons.wikimedia.org/wiki/File:Old\_paper3.jpg}
    \item Old paper 4:
        \url{https://commons.wikimedia.org/wiki/File:Old\_paper4.jpg}
    \item Old paper 6:
        \url{https://commons.wikimedia.org/wiki/File:Old\_paper6.jpg}
    \item Old paper 7:
        \url{https://commons.wikimedia.org/wiki/File:Old\_paper7.jpg}
    \item Vinatage Paper Texture:
        \href{https://commons.wikimedia.org/wiki/File:Vintage\_Paper\_Texture\_(9789792113).jpg}
        {https://commons.wikimedia.org/wiki/File:\\Vintage\_Paper\_Texture\_(9789792113).jpg}
    \item Blank page, brown paper texture:
        \url{https://commons.wikimedia.org/wiki/File:Blank\_page,\_brown\_paper\_texture\_(14802136533).jpg}
    \item Parchment 00:
        \url{https://commons.wikimedia.org/wiki/File:Parchment.00.jpg}
\end{itemize}
Besides these more authentic paper backgrounds, some artists' free-to-use interpretations of antique paper were also used:
\begin{itemize}
    \item Pixabay empty brown canvas on Pexels:
        \url{https://www.pexels.com/photo/abstract-ancient-antique-art-235985/}
    \item HD paper texture by Imrooniel on Deviant Art: \url{https://www.deviantart.com/imrooniel/art/HD-paper-texture-298160595}
    \item 5 paper textures by MarshmellowHeaven that were stained with coffee and tea:
        \begin{itemize}
            \item \href{https://www.deviantart.com/marshmellowheaven/art/Texture-Paper-1-195235719}
                {https://www.deviantart.com/marshmellowheaven/art/Texture-Paper-1-195235719}
            \item \href{https://www.deviantart.com/marshmellowheaven/art/Texture-Paper-2-195236191}
                {https://www.deviantart.com/marshmellowheaven/art/Texture-Paper-2-195236191}
            \item \href{https://www.deviantart.com/marshmellowheaven/art/Texture-Paper-3-195236939}
                {https://www.deviantart.com/marshmellowheaven/art/Texture-Paper-3-195236939}
            \item \href{https://www.deviantart.com/marshmellowheaven/art/Texture-Paper-4-195237220}
                {https://www.deviantart.com/marshmellowheaven/art/Texture-Paper-4-195237220}
            \item \href{https://www.deviantart.com/marshmellowheaven/art/Texture-Paper-5-195237843}
                {https://www.deviantart.com/marshmellowheaven/art/Texture-Paper-5-195237843}
        \end{itemize}
\end{itemize}
These background images were randomly selected for every handwriting line image that was chosen to be of the Antique class.
All backgrounds were turned to grayscale, while remaining in RGB color space to match IAM and RIMES formats.
Given the chosen background image, a cropping of that background that fit the size of the handwriting line image was selected and the handwriting line was laid over that cropped background.
The exact procedure is available in the provided code, which will be publicly available upon publication.
While the Large-Scale evaluation did not assess ODAI as the 5-fold CV experiments did in the main paper, it did use the same code with different backgrounds to assess how the baseline agents performed when the background changed.

\begin{table*}[t]
    \centering
    \scalebox{0.84}{{\setlength{\tabcolsep}{0.4em}
    \begin{tabular}{l|ll|ll} 
        \multicolumn{5}{c}{\textbf{Training Set Mean Measures of 5-fold Cross Validation}} \\ 
        \hline
        \textbf{Task} & \multicolumn{2}{c}{\textbf{Multi-class Classif.}} & \multicolumn{2}{|c}{\textbf{Binary Novelty}} \\ 
         & \multicolumn{2}{c}{\textbf{with Novel Class}} & \multicolumn{2}{|c}{\textbf{Detection}} \\ 
        \multicolumn{1}{r|}{\textbf{Model}} & \textbf{NMI} & \textbf{Acc.} & \textbf{NMI} & \textbf{Acc.} \\ 
        \hline
        \textbf{Writer ID} & &  \\
        \multicolumn{1}{r|}{Mean HOG EVM}        & 0.8198 $\pm$ 2.09e-3 & 0.8444 $\pm$ 8.58e-4  & 0.9557 $\pm$ 1.12e-3 & 0.9889 $\pm$ 3.38e-4 \\
        \multicolumn{1}{r|}{10-Mean HOG EVM}     & 0.9871 $\pm$ 9.55e-4 & 0.9921 $\pm$ 4.37e-3  & 0.9586 $\pm$ 3.35e-3 & 0.9900 $\pm$ 9.38e-4 \\
        \multicolumn{1}{r|}{ResNet50 EVM}        & 1.0 $\pm$ 0.0 & 1.0 $\pm$ 0.0    & 1.0 $\pm$ 0.0 & 1.0 $\pm$ 0.0 \\
        \multicolumn{1}{r|}{CRNN-PCA EVM}        & 0.9996 $\pm$ 9.90e-5 & 0.9998 $\pm$ 4.07e-3  & 0.9990 $\pm$ 3.94e-4 & 0.9998 $\pm$ 6.18e-5 \\
        \hline
        \textbf{Appearances (ODAI)} &  & \\
        \multicolumn{1}{r|}{Mean HOG EVM}        & 0.6611 $\pm$ 2.47e-3 & 0.8559 $\pm$ 1.46e-3  & 0.4817 $\pm$ 4.04e-3 & 0.7654 $\pm$ 2.78e-3  \\
        \multicolumn{1}{r|}{10-Mean HOG EVM}     & 0.7124 $\pm$ 1.33e-3 & 0.8973 $\pm$ 5.74e-4  & 0.6076 $\pm$ 3.02e-3 & 0.8348 $\pm$ 1.74e-3 \\
        \multicolumn{1}{r|}{ResNet50 EVM}        & 0.8327 $\pm$ 5.44e-6 & 0.8000 $\pm$ 4.61e-6  & 0.4326 $\pm$ 6.93e-6 & 0.6667 $\pm$ 6.40e-6 \\
        \multicolumn{1}{r|}{CRNN-PCA EVM}        & 0.4455 $\pm$ 3.19e-3 & 0.6546 $\pm$ 5.27e-3  & 0.0840 $\pm$ 6.78e-3 & 0.3089 $\pm$ 1.31e-2 \\
        \hline
        & \textbf{Character Acc.} & \textbf{Word Acc.} & \textbf{NMI}& \textbf{Acc.} \\ 
        \hline
        \textbf{Transcription} & &  \\
        \multicolumn{1}{r|}{CRNN} & 0.9904 $\pm$ 5.83e-4 & 0.9660 $\pm$ 1.98e-3 & 0.9601 $\pm$ 3.17e-3 & 0.9913 $\pm$ 8.07e-4 \\
    \end{tabular}
    }}
    \caption{
    The mean 5-fold results with standard error for the train split of all three experiments.
    ``NMI" stands for Normalized Mutual Information.
    All measures reported here are found after selecting the maximum probable class as predicted by the classifier after thresholding the maximum probability to determine if novel.
    }
    \label{tab:mean_res_train}
\end{table*}
\begin{table*}[t]
    \centering
    \scalebox{0.84}{{\setlength{\tabcolsep}{0.4em}
    \begin{tabular}{l|ll|ll} 
        \multicolumn{5}{c}{\textbf{Validation Set Mean Measures of 5-fold Cross Validation}} \\ 
        \hline
        \textbf{Task} & \multicolumn{2}{c}{\textbf{Multi-class Classif.}} & \multicolumn{2}{|c}{\textbf{Binary Novelty}} \\ 
         & \multicolumn{2}{c}{\textbf{with Novel Class}} & \multicolumn{2}{|c}{\textbf{Detection}} \\ 
        \multicolumn{1}{r|}{\textbf{Model}} & \textbf{NMI} & \textbf{Acc.} & \textbf{NMI} & \textbf{Acc.} \\ 
        \hline
        \textbf{Writer ID} & &  \\
        \multicolumn{1}{r|}{Mean HOG EVM}        & 0.7394 $\pm$ 9.28e-3 & 0.7123 $\pm$ 4.93e-3  & 0.7754 $\pm$ 1.66e-2 & 0.9265 $\pm$ 7.19e-3 \\
        \multicolumn{1}{r|}{10-Mean HOG EVM}     & 0.6497 $\pm$ 7.90e-3 & 0.7852 $\pm$ 3.83e-3    & 0.3857 $\pm$ 8.17e-3 & 0.6246 $\pm$ 6.13e-3 \\
        \multicolumn{1}{r|}{ResNet50 EVM}        & 0.6403 $\pm$ 8.56e-3 & 0.7876 $\pm$ 3.23e-3    & 0.3793 $\pm$ 6.18e-3 & 0.6126 $\pm$ 6.18e-3 \\
        \multicolumn{1}{r|}{CRNN-PCA EVM}        & 0.6513 $\pm$ 7.95e-3 & 0.8074 $\pm$ 3.54e-3   & 0.3949 $\pm$ 6.96e-3 & 0.6266 $\pm$ 6.77e-3 \\
        \hline
        \textbf{Appearances (ODAI)} &  & \\
        \multicolumn{1}{r|}{Mean HOG EVM}        & 0.5809 $\pm$ 2.54e-3 & 0.7886 $\pm$ 1.96e-3  & 0.3358 $\pm$ 3.86e-3 & 0.6464 $\pm$ 3.69e-3  \\
        \multicolumn{1}{r|}{10-Mean HOG EVM}     & 0.4948 $\pm$ 4.14e-3 & 0.7525 $\pm$ 1.96e-3  & 0.2894 $\pm$ 2.75e-3 & 0.5799 $\pm$ 2.49e-3 \\
        \multicolumn{1}{r|}{ResNet50 EVM}        & 0.0272 $\pm$ 1.18e-3 & 0.5097 $\pm$ 4.11e-4  & 0.0181 $\pm$ 7.61e-4 & 0.0989 $\pm$ 2.21e-3 \\
        \multicolumn{1}{r|}{CRNN-PCA EVM}        & 0.0177 $\pm$ 1.87e-3 & 0.4315 $\pm$ 5.96e-3  & 0.0027 $\pm$ 1.42e-2 & 0.4848 $\pm$ 6.28e-3 \\
        \hline
        & \textbf{Character Acc.} & \textbf{Word Acc.} & \textbf{NMI} & \textbf{Acc.} \\ 
        \hline
        \textbf{Transcription} & &  \\
        \multicolumn{1}{r|}{CRNN} & 0.9516 $\pm$ 3.53e-3 & 0.8861 $\pm$ 2.61e-3 & 0.8787 $\pm$ 7.03e-3 & 0.9664 $\pm$ 2.44e-3 \\
    \end{tabular}
    }}
    \caption{
    The mean 5-fold results with standard error for the validation split of all three experiments.
    ``NMI" stands for Normalized Mutual Information.
    All measures reported here are found after selecting the maximum probable class as predicted by the classifier after thresholding the maximum probability to determine if novel.
    }
    \label{tab:mean_res_val}
\end{table*}

\section{Large-Scale 55K Test Evaluation: Additional Protocol Information and Detailed Analysis}

The Mean HOG configuration of the baseline open world HWR agent was evaluated with 55,000 tests.  We generate 5,500 tests based on experimental conditions.  For each generated test, we create nine additional tests, re-ordering the test samples to average-out sample variations while retaining the same conditions of the test.  Tests were constructed and grouped by types of novelty. The tests were constructed to evaluate both single writing sample novelty detection and world change detection indicated by data distribution change from a non-novelty phase to a novelty phase of the test. In addition to establishing a foundation for novelty detection and characterization in HWR, in this evaluation, we establish some initial metrics for novelty difficulty, identifying factors impacting the performance of detecting novelty and transcribing handwritten text. 

\subsection{Protocol: Modified IAM Off-Line Handwriting Data.}
The roughly 55,000 novel writing samples used in evaluation were constructed from modified samples of the IAM Offline Handwriting Dataset~\cite{marti2002iam}. The training data will be publicly released after this paper's publication.   A representative portion of the tests will be released as well.

Training and evaluation data, in the form of individual lines, was selected from IAM.  Prior to training, lines were denoised, removing shadow boxes around the letters of each word.  Features were then extracted from the clean lines of written text to capture writing characteristics including pen pressure, letter slant, word spacing and character size~\cite{joshi2015}. A distance matrix was formed between by the sum of absolute differences between each writer's mean style across all example words from each writer. The distance matrix served as a writer similarity measurement.

The training set was made up lines of text from 50 selected writers representing a subset of the writer style descriptor values, leaving one or two bins for each feature excluded for use in the novelty evaluation set. Lines of text did not contain any additional effects, using a white background.  Sample lines from six additional writers were chosen to compose an unknown writer training set. The set was supplemented with samples from the RIMES dataset and samples from the same 50 writers with background effects including salt and pepper noise,  antique paper, and faded impressions of shaded boxes around the words in each line of text. 

The evaluation set was made up of the remaining writers and writing sample manipulations to alter characteristics of both the writing style and the background.   Letter style manipulations included thinning or widening, brightness, resizing  and slant adjustments to each line of text.  The background was composed from Creative Commons licensed images of textured paper.  Pen manipulations were similarly constructed by merging in textures and colors, weighted by the pixel strength (\textit{i.e.}, pen pressure). 

The difficulty associated with each test is determined by the novelty type.  The difficulty for novel writers and novel letter manipulations was determined by the ontological separation of four writing style features: pen pressure, letter slant, character size, and word spacing.  Grouping novel writers with non-novel writers with similar styles is intended to make detection more difficult.  The difficulty of novel pen and backgrounds was measured by the inverse intensity of the background (since the letters are black). 

Most novel examples were constructed with a single type of novelty.  Background and pen novelties were applied to sample text lines from the 50 known writers.   The number of text lines per test varied based on availability of data targeting the specific novelty:  512, 768, or 1,024.  In total, each test selected from 1,696 non-novel examples of the 50 known writers and approximately 50,000 novelties. Tests were composed of writing samples selected and organized by six independent discrete variables defining the experimental conditions of each test to explore the performance regime in novelty detection, resulting in 3,888 unique combinations. Using several subtypes (\textit{e.g.}, different backgrounds) of novelties by type and difficulty, we constructed approximately 5,500 tests, each reordered nine times to average out sample variations.

Difficulty and novelty type (Table~\ref{tab:novety_instance_count}) affect writer prediction and transcription accuracy.  Variables (Table~\ref{tab:novel_vars}) associated with distribution and placement of novelty in stream of data, such as introduction point, density of novel to non-novel samples and distribution type are varied to measure impact on novelty detection.

\begin{table}[h]
\centering
  \begin{tabular}{|l r|}
  \hline
  \textbf{Novelty Type} & \textbf{Count} \\
  \hline
    Background   &  17,662 \\
      \hline
    Letter       &  11,868 \\
      \hline
    Pen          &  11,289 \\
      \hline
    Writer       &  8,427 \\
      \hline
    No Novelty    &  1,696 \\
      \hline
\end{tabular}
    \caption{Number of writing samples for each type of novelty.}
    \label{tab:novety_instance_count}
\end{table}

\begin{table}
    \centering
    \begin{tabular}{|l l|} 
        \hline
        \textbf{Independent Variables} & \textbf{Values}  \\
        \hline
        Mean Novelty  & 0.4, 0.5, 0.6, 0.7, 0.8, 0.9 \\ 
        Introduction Pt. & \\
        \hline
        Density of Novelty & 6 different densities \\
        \hline
        Novelty Type & Writer, Letter, Background \\
        \hline
        Difficulty & Easy, Medium, Hard \\
        \hline
        Distribution Type & High (positive skew),\\
        & Low (negative skew),\\
        & Mid (normal), Flat (uniform) \\
        \hline
        Test Length & 512, 768 and 1,024 \\
        \hline
    \end{tabular}
    \caption{Independent Variables forming the experimental conditions of each Novelty Test}
    \label{tab:novel_vars}
\end{table}

\subsection {Supplemental Results for the Large-Scale 55K Evaluation}
\label{sec:supp_eval_results}

This section provides a more fine-grained analysis over the 55,000 tests presented to the closed world agents and the novelty detecting open world agent described in the main text of the paper. The agent configuration used for the open world agent experiments utilizes HOG features for all style tasks. For this analysis, we present general novelty detection, text transcription and writer identification performance across all tests based on types of novelty.

\subsubsection{Closed World Agents: Transcription and Writer Identification}

As expected, novelty negatively impacted the writer identification and sample transcription accuracy.
Results are shown in Table~\ref{tab:baseline_results}.
Mean Character Transcription Accuracy is reported as ${1-L(G_s,A_s)/\max(|{G_s}|,|{A_s}|)}$ where ${L}$ is Levenshtein Edit Distance, ${G_s}$ is ground truth text for writing sample ${s}$, and ${A_s}$ is the agent's predicted transcription for writing sample ${s}$, averaged over the ten variations of each test.
Writer Identification Accuracy is reported as mean accuracy of the top-1 and top-3 predictions out of $K$+1 writers, where $K$ = 50 for all tests, and the additional class is for novel writers. 

\begin{table}[h]
\centering
  \begin{tabular}{|l |c c c |}
      \hline
       & \textbf{Mean} & \textbf{Writer ID} & \textbf{Writer ID}\\
      \textbf{Is Novel?} &
      \textbf{Char. Acc.} & 
      \textbf{Top-3 Acc.} & \textbf{Top-1 Acc.} \\
      \hline
       False & 0.85 & 0.99 & 0.99 \\
      \hline
       True & 0.47 & 0.40 & 0.24 \\
      \hline
   \end{tabular}
   \caption{Baseline closed world agent mean character transcription accuracy, top-3 writer identification accuracy, and top-1 writer identification accuracy in response to non-novel and novel writing samples. }
 \label{tab:baseline_results}
\end{table}

\subsubsection{Open World Agent: Novel vs. Non-Novel Predictions}
Again as expected, novelty negatively impacted both text transcription and writer identification accuracy. However, the open world agent is significantly better at the text transcription task. Results are shown in Table~\ref{tab:agent_results}. Transcription performance is reported as mean character accuracy computed using the ground truth and the agent provided transcriptions for all tests. Writer identification accuracy is reported as mean accuracy of the Top-1 and Top-3 predictions out of $K$+1 writers, where $K$ = 50 for all tests. 

\begin{table}[h]
\centering
  \begin{tabular}{|l | c c c |}
      \hline
      & \textbf{Mean} & \textbf{Writer ID} & \textbf{Writer ID} \\
      \textbf{Is Novel?} & \textbf{Char. Acc.} & \textbf{Top-3 Acc.} & \textbf{Top-1 Acc.}  \\
      \hline
       False &  0.82 &  0.942 &  0.719 \\
       \hline
       True  & 0.62 &  0.479 &  0.220 \\
       \hline
   \end{tabular}
   \caption{Baseline open world agent mean character transcription accuracy, top-3 writer identification accuracy, and top-1 writer identification accuracy in response to non-novel and novel writing samples.}
 \label{tab:agent_results}
\end{table}

\subsubsection{Open World Agent: Novel Style Manipulations}
Style manipulations include manipulations to the characters.  These manipulations had a measurable impact on writer identification performance. Results are shown in Table~\ref{tab:novel_style}.  Dilating the letters did not affect performance, down-weighting pen width as a major factor of a writer's style.  More extreme character manipulations such as large slants and slants coupled with dilation were more easily detected as being novel, as expected. Inverting pixel values for written text did not adversely affect writer identification performance.  The novelty detector did not equate letter inversion as novelty.  Each novelty type was represented by 1,696 sample images.

Four different summary statistics are computed for the novel style manipulations.
Novelty Detection Accuracy is mean accuracy of all of the detection decisions.
Mean Character Transcription Accuracy is defined as

\begin{equation}
{1-L(G_s,A_s)/\max(|{G_s}|,|{A_s}|)}
\end{equation}
where ${L}$ is Levenshtein Edit Distance, ${G_s}$ is ground truth text for writing sample ${s}$, ${A_s}$ is the agent's predicted transcription for writing sample ${s}$, averaged over the ten variations of each test.
NMI represents normalized mutual information between the actual writer of the sample and the top-1 predicted writer.
Writer Identification Accuracy is mean accuracy of the top-3 predictions out of $K$+1 writers, where $K$ = 50 across all tests, and the additional class is for novel writers.
These summary statistics are also used for the novel pens and novel backgrounds assessments, which are described below. 

\begin{table}[h]
\centering
  \begin{tabular}{|l | c c c c|} 
   \hline
   & \multicolumn{1}{c}{\textbf{Novelty}} & \multicolumn{1}{c}{\textbf{Mean}} && \multicolumn{1}{c|}{\textbf{Writer}} \\
   \textbf{Novelty} & \multicolumn{1}{c}{\textbf{Detection}} &  \multicolumn{1}{c}{\textbf{Char.}} & &  \multicolumn{1}{c|}{\textbf{ID}} \\
	 \textbf{Type} & \multicolumn{1}{c}{\textbf{Acc.}} &  \multicolumn{1}{c}{\textbf{Acc.}} &  \multicolumn{1}{c}{\textbf{NMI}} &  \multicolumn{1}{c|}{\textbf{Acc.}} \\
   \hline
    Dilate          & 0.99 & 0.70 & 0.01 & 0.57 \\
   \hline
   Erode            & 0.79 & 0.77 & 0.35 & 0.68  \\
   \hline
   Increase Size    & 0.99 & 0.33 & 0.01 & 0.02 \\
   \hline
   Big Right Slant  & 0.79 & 0.62 & 0.21 & 0.09  \\
   \hline
   Slant w/ Dilate  & 0.99 & 0.46 & 0.04 & 0.00 \\
   \hline
   Big Left Slant   & 1.00 & 0.55 & 0.01 & 0.02 \\
   \hline
   Small Slant      & 0.86 & 0.52 & 0.23 & 0.04 \\
   \hline
   Inverted         & 0.33 & 0.71 & 0.79 & 0.94 \\
   \hline
    \end{tabular}
    \caption{Novelty detection accuracy, mean character transcription accuracy, top-1 writer identification mean normalized mutual information, and top-3 writer identification accuracy given pen novelties grouped by novel style changes.}
    \label{tab:novel_style}
\end{table}

\subsubsection{Open World Agent: Novel Pens}

Novel Pens include manipulations to written text, replacing the pixels with textures and colors, weighted by the intensity of the pen as described by pen pressure. Results are shown in Table~\ref{tab:novel_pen}.  Pen manipulations had minimal impact on writer identification performance.
Each novelty type was represented by 1,696 sample images.

\begin{table}[h]
\centering
 \begin{tabular}{| l | c c c c |} 
 \hline
   & \multicolumn{1}{c}{\textbf{Novelty}} & \multicolumn{1}{c}{\textbf{Mean}} && \multicolumn{1}{c|}{\textbf{Writer}} \\
   \textbf{Novelty} & \multicolumn{1}{c}{\textbf{Detection}} &  \multicolumn{1}{c}{\textbf{Char.}} & &  \multicolumn{1}{c|}{\textbf{ID}} \\
	 \textbf{Type} & \multicolumn{1}{c}{\textbf{Acc.}} &  \multicolumn{1}{c}{\textbf{Acc.}} &  \multicolumn{1}{c}{\textbf{NMI}} &  \multicolumn{1}{c|}{\textbf{Acc.}} \\
   \hline
Blue Color      & 0.98 & 0.78 & 0.09 & 0.53 \\
 \hline
Brown Texture   & 0.74 & 0.75 & 0.43 & 0.79 \\
 \hline
Gold Texture    & 0.92 & 0.69 & 0.22 & 0.73 \\
 \hline
Rainbow         & 0.98 & 0.71 & 0.10 & 0.57 \\
 \hline
Red Color       & 0.98 & 0.70 & 0.10 & 0.53 \\
 \hline
\end{tabular}
\caption{Novelty detection accuracy, mean character transcription accuracy, top-1 writer identification mean normalized mutual information, and top-3 writer identification accuracy grouped by novel pens.}
\label{tab:novel_pen}
\end{table}

\subsubsection{Open World Agent: Novel Backgrounds}

Background manipulation had a more diverse impact on writer prediction performance than style manipulations. Results are shown in Table~\ref{tab:novel_backgrounds}.  NMI represents normalized mutual information between actual sample writer and top-1 predicted writer.  Novel types of shadow boxes (from the uncleaned lines extracted from IAM) had the highest writer identification accuracy, perhaps due to similar associations made with these types of artificial irregularities in the training set.  As with pen manipulations, more extreme manipulations resulted in higher detection accuracy. Increased texture interfered with the agent's ability to identify the writer.  

\begin{table}[h]
\center
\begin{tabular}{| l | c c c c |}
 \hline
   & \multicolumn{1}{c}{\textbf{Novelty}} & \multicolumn{1}{c}{\textbf{Mean}} && \multicolumn{1}{c|}{\textbf{Writer}}\\ 
   \textbf{Novelty} & \multicolumn{1}{c}{\textbf{Detection}} &  \multicolumn{1}{c}{\textbf{Char.}} & &  \multicolumn{1}{c|}{\textbf{ID}} \\
	 \textbf{Type} & \multicolumn{1}{c}{\textbf{Acc.}} &  \multicolumn{1}{c}{\textbf{Acc.}} &  \multicolumn{1}{c}{\textbf{NMI}} &  \multicolumn{1}{c|}{\textbf{Acc.}} \\
 \hline
Antique        & 0.40 & 0.80 & 0.47 & 0.39 \\
 \hline
Blue Fabric    & 0.98 & 0.40 & 0.10 & 0.42 \\
 \hline
Blue Color     & 0.98 & 0.69 & 0.01 & 0.54 \\
 \hline
Blue Wall      & 0.99 & 0.33 & 0.01 & 0.10 \\
 \hline
Brown Fabric   & 1.00 & 0.60 & 0.01 & 0.11 \\
 \hline
Crinked Paper  & 1.00 & 0.71 & 0.02 & 0.16 \\
 \hline
Gaussian Noise & 1.00 & 0.14 & 0.00 & 0.16 \\
 \hline
Gold Wall      & 0.68 & 0.51 & 0.34 & 0.37 \\
 \hline
Rainbow Paper  & 0.98 & 0.25 & 0.12 & 0.54 \\
 \hline
Shadow Boxes   & 0.59 & 0.88 & 0.62 & 0.91 \\
 \hline
\end{tabular}
\caption{Novelty detection, mean character transcription accuracy, top-1 writer identification mean normalized mutual information, and top-3 novel writer identification accuracy grouped by writing style.}
\label{tab:novel_backgrounds}
\end{table}

\subsubsection {Open World Agent: Writer Similarity in Novel Writer Discovery}

Each test is composed of sample writing from known and unknown writers.  Here we find the minimum distance of an unknown writer across all known writers.  We hypothesize that the greater the distance of writer style attributes of unknown writers with known writers, as captured in the ontological specification, the easier it is to detect a novel writer.

Surprisingly, the results did not show a strong correlation as expected.
Table~\ref{tab:det_pred_acc} shows the Pearson's correlation of each style attribute with detection and top-1 novel writer identification accuracy.
We believe this due to two key factors: not enough variability in writing styles in the unknown population and the chosen set of attributes insufficiently capturing all of the essential characteristics of writing style.  Pen pressure had the highest correlation of the four ontological specified factors.  Collectively, a weak positive correlation did support the hypothesize. The proposed benchmark can be augmented with additional attributes, as the challenge problem evolves.  

\begin{table}[h]
\centering
\begin{tabular}{| l | c c |}
\hline
\textbf{Style} & \shortstack{\textbf{Novelty} \\ \textbf{Det. Corr.}} & \shortstack{\textbf{Writer} \\ \textbf{ID Corr.}} \\
 \hline
Slant Angle  &  0.06 &  0.01 \\
 \hline
Skew Angle   &  0.02 &  0.04 \\
 \hline
Word Spacing & -0.02 & -0.05 \\
 \hline
Pen Pressure &  0.14 &  0.09 \\
 \hline
Character Size  &  0.03 &  0.04 \\
 \hline
Summed       &  0.12 &	0.18 \\
 \hline
\end{tabular}
\caption{Novelty detection and novel writer identification correlation grouped by writing style.}
\label{tab:det_pred_acc}
\end{table}

\subsubsection{Open World Agent: Factors in Novelty Detection}

\begin{figure}[h!]
    \centering
  \includegraphics[scale=0.6]{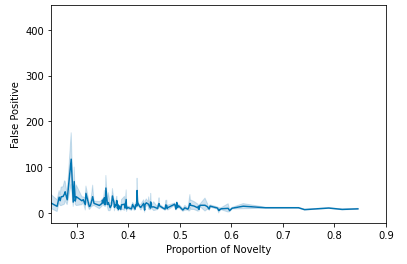}
  \caption{False positive count by the proportion of novelty present.}
  \label{fig:fp_prop_novelty}
\end{figure}

A critical factor in the 55K tests is the density and location of novelty introduction --- the switch between pre-novelty and post-novelty phases of the test given a stream of writing samples.  This approach treats novelty as perceived world changing events rather than outliers, where confidence of novelty predictions increases as more novel examples are encountered in the data stream, increasing the body of evidence.   With this approach, the level of false positives, those misidentified non-novel examples that fall in the pre-novelty phase of the test, can be substantially reduced. Fig.~\ref{fig:fp_prop_novelty} shows the false positive count by the proportion of novelty.  The variability and amount of false positives decreases as the proportion of novel samples to non-novel samples increases.

We conducted ANOVA to identify factors affecting the false positive rate (see Table \ref{tab:anova}). Along with the proportion of novelty, distribution type had a significant impact on the false positive rate. A positively skewed distribution, where novel samples densely occur at the start of the novelty phase of the test, is associated with lower false positive rate when compared to other distribution types such as a negatively skewed distribution.  
Novelty difficulty had a weak association to the false positive rate.

\begin{table}[h]
\centering
\begin{tabular}{| l | r r r r|}
\hline
\textbf{Factor} & Sum of Squares & $df$ & F & $p$ \\
 \hline
 Distribution   &&&& \\
 Type  &  1.058e+06 &  3 & 129.300 & 0.000 \\
 \hline
Level of &&&& \\
Difficulty &  7.456e+03 &  2 & 1.365 & 0.255  \\
 \hline
Location of &&&& \\
 Novelty & 2.646e+06 &  1 & 969.301 & 0.000  \\
 \hline
Proportion of &&&& \\
Novelty & 5.848e+05 &  1 & 214.240 & 0.000  \\
 \hline
Residual  & 1.205e+00 &  44170 &  &  \\
 \hline
\end{tabular}
\caption{ANOVA analysis of statistical influence given several test generating independent variables identified in Table \ref{tab:novel_vars} on false positive rate.}
\label{tab:anova}
\end{table}

\bibliographystyle{splncs04}
\bibliography{bibs/background, bibs/data, bibs/theory}